\documentclass[10pt, conference, compsocconf]{IEEEtran}
\usepackage{ifpdf}
\usepackage{cite}
\ifCLASSINFOpdf
\usepackage[pdftex]{graphicx}
\else
\usepackage[dvips]{graphicx}
\fi
\usepackage[cmex10]{amsmath}
\usepackage{algorithmic}
\usepackage{array}
\usepackage{mdwmath}
\usepackage{mdwtab}
\usepackage{eqparbox}
\usepackage{epsfig}
\usepackage[most]{tcolorbox}
\usepackage[tight,footnotesize]{subfigure}
\usepackage{caption}
\usepackage[font=footnotesize]{subfig}
\usepackage{url}
\usepackage{soul}

\begin{document}
\title{Graphical Object Detection in Document Images}
\author{\IEEEauthorblockN{Ranajit Saha, Ajoy Mondal and C V Jawahar}
\IEEEauthorblockA{Centre for Visual Information Technology,\\
International Institute of Information Technology, Hyderabad, India, \\
Email: ranajit.saha@students.iiit.ac.in, ajoy.mondal@iiit.ac.in and jawahar@iiit.ac.in}}
\maketitle

\begin{abstract}
Graphical elements: particularly tables and figures contain a visual summary of the most valuable information contained in a document. Therefore, localization of such graphical objects in the document images is the initial step to understand the content of such graphical objects or document images. In this paper, we present a novel end-to-end trainable deep learning based framework to localize graphical objects in the document images called as Graphical Object Detection (\textsc{god}). Our framework is data-driven and does not require any heuristics or meta-data to locate graphical objects in the document images. The \textsc{god} explores the concept of transfer learning and domain adaptation to handle scarcity of labeled training images for graphical object detection task in the document images. Performance analysis carried out on the various public benchmark data sets: {\sc \textbf{icdar-2013}}, {\sc \textbf{icdar-pod 2017}} and {\sc \textbf{unlv}} shows that our model yields promising results as compared to state-of-the-art techniques.
\end{abstract}


\IEEEpeerreviewmaketitle

\section{Introduction}\label{introduction}

With the rapidly increasing number of digital documents, the manual extraction and retrieval of information from them has become an in-feasible option - Automated methods are coming to the rescue. There are many tools and methods available to convert the digital documents into process-able texts. The graphical page elements, such as tables, figures, equations play a very important role in understanding and extracting the information from the documents. Therefore, the detection of those objects from the documents has attracted a lot of attention in the research community. Because of no common dimensions and diversity of layouts of the tables and figures in the documents, detection of them is considered to be a challenging problem.

\begin{figure}[ht!]
\centerline{
\tcbox[sharp corners, size = tight, boxrule=0.2mm, 
            colframe=black, colback=white]{
\psfig{figure=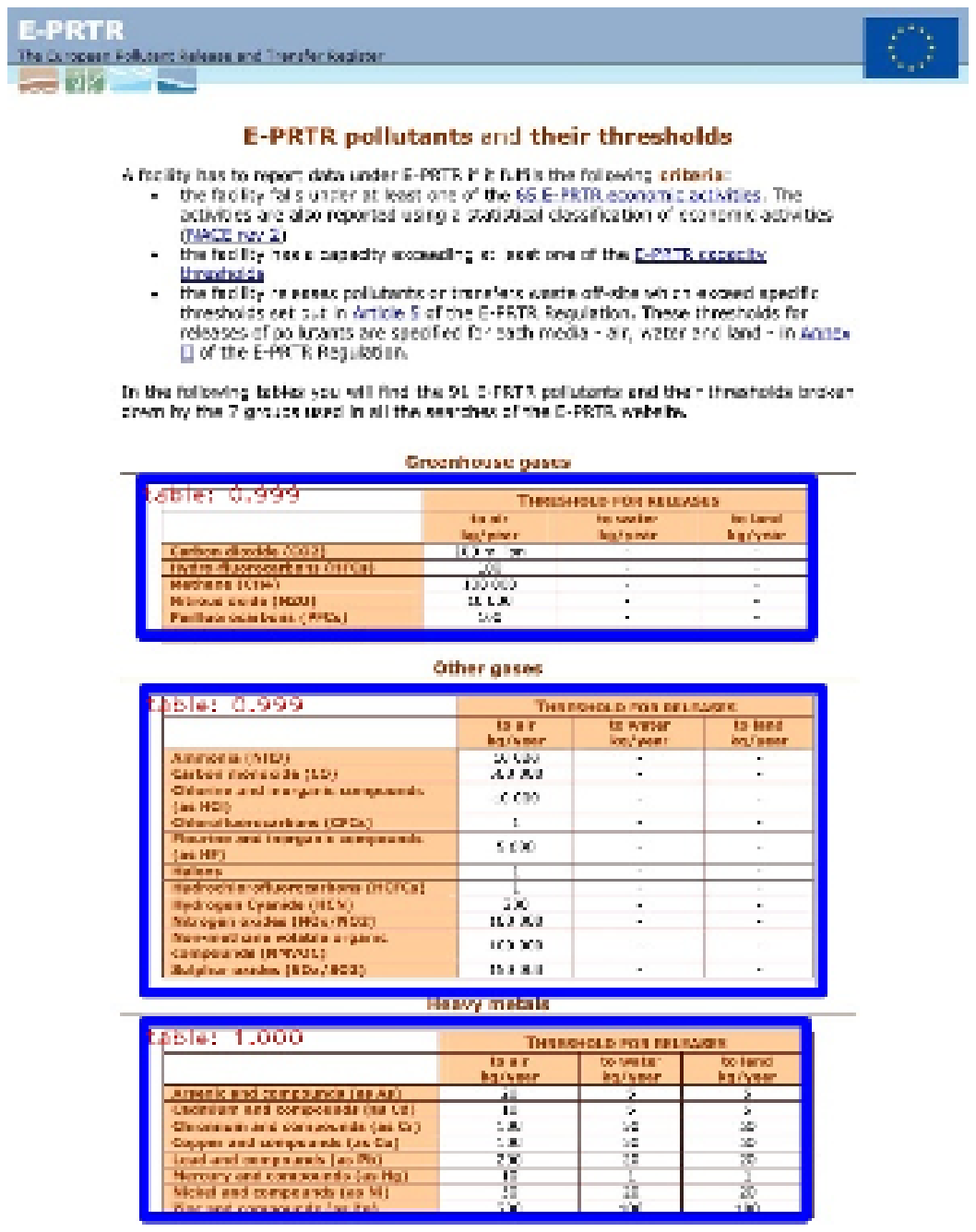, width=0.11\textwidth,height=0.14\textwidth}}
\hspace{-0.01\textwidth}
\tcbox[sharp corners, size = tight, boxrule=0.2mm, colframe=black, colback=white]{
\psfig{figure=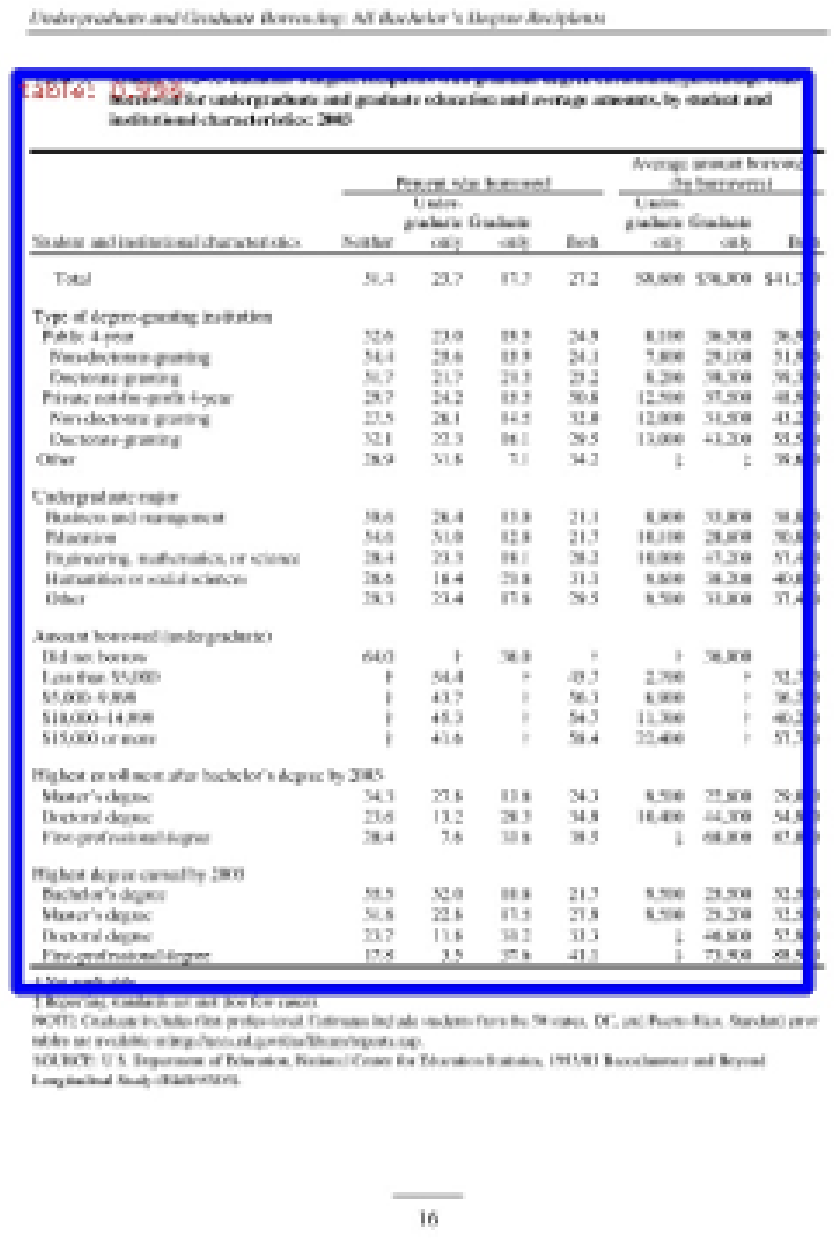, width=0.11\textwidth,height=0.14\textwidth}}
\hspace{-0.01\textwidth}
\tcbox[sharp corners, size = tight, boxrule=0.2mm, colframe=black, colback=white]{
\psfig{figure=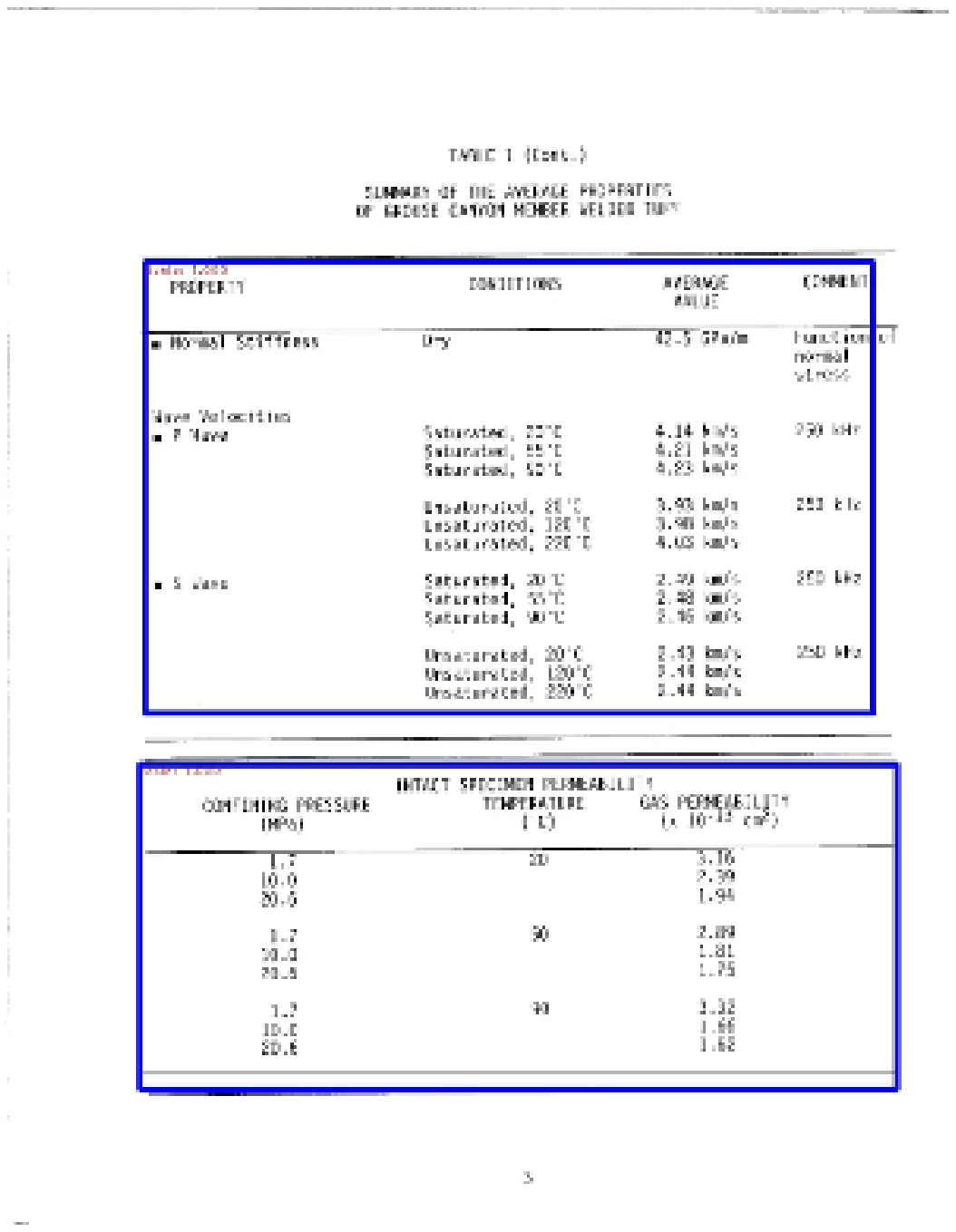, width=0.11\textwidth,height=0.14\textwidth}}
\tcbox[sharp corners, size = tight, boxrule=0.2mm, colframe=black, colback=white]{
\psfig{figure=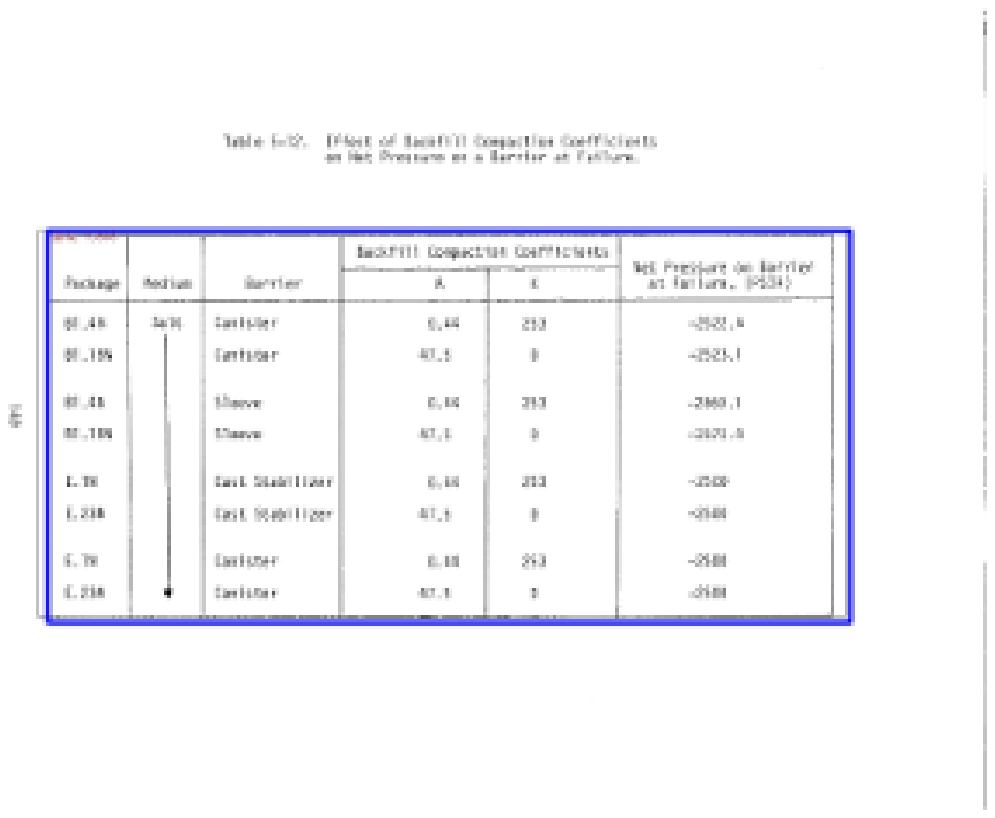, width=0.11\textwidth,height=0.14\textwidth}}}
\vspace{-0.02\textwidth}
\centerline{
\tcbox[sharp corners, size = tight, boxrule=0.2mm,colframe=black, colback=white]{
\psfig{figure=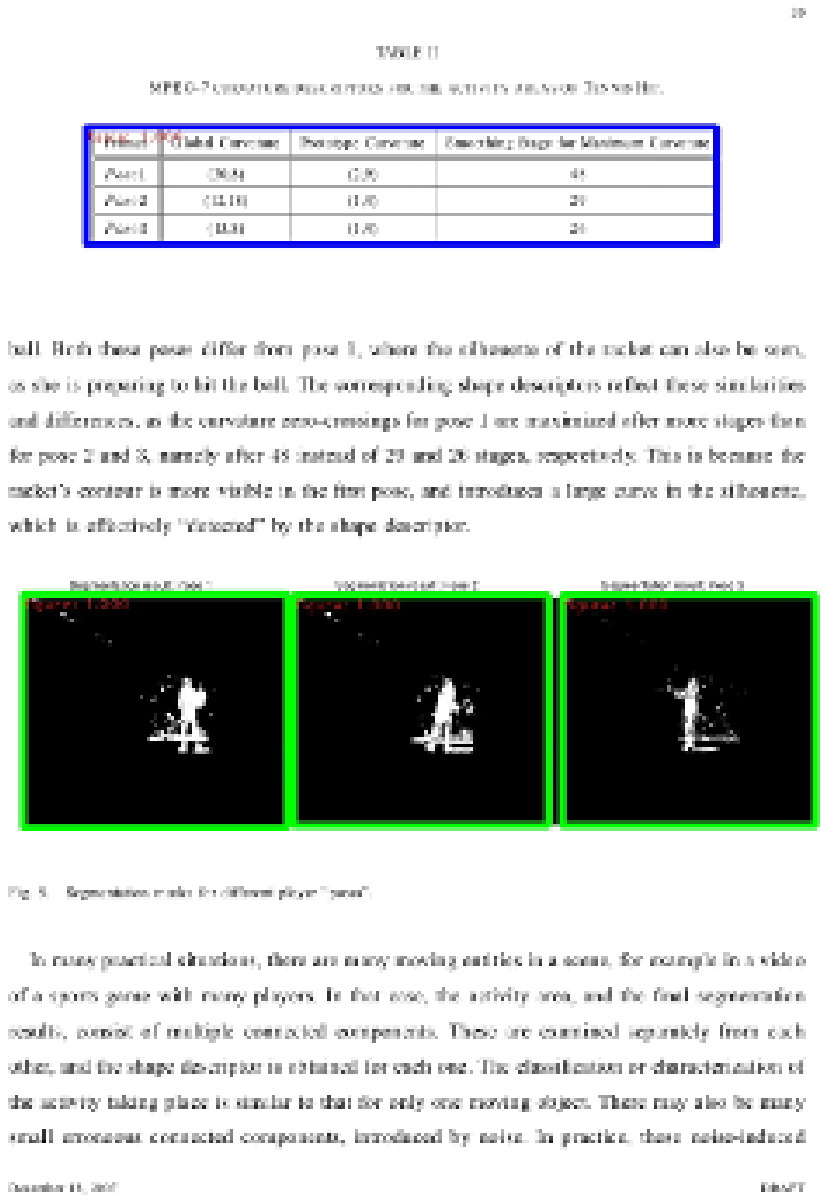, width=0.11\textwidth,height=0.14\textwidth}}
\hspace{-0.01\textwidth}
\tcbox[sharp corners, size = tight, boxrule=0.2mm, colframe=black, colback=white]{
\psfig{figure=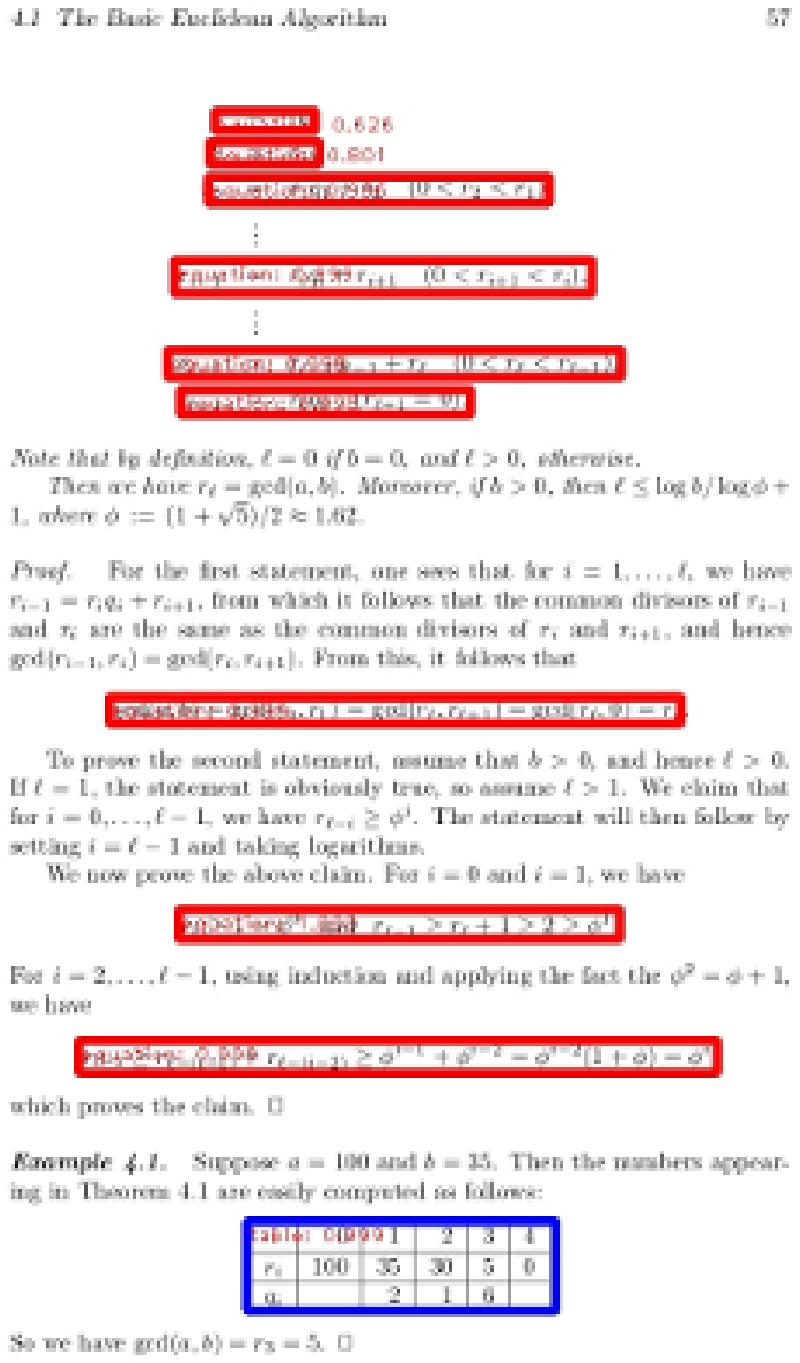, width=0.11\textwidth,height=0.14\textwidth}}
\hspace{-0.01\textwidth}
\tcbox[sharp corners, size = tight, boxrule=0.2mm, colframe=black, colback=white]{
\psfig{figure=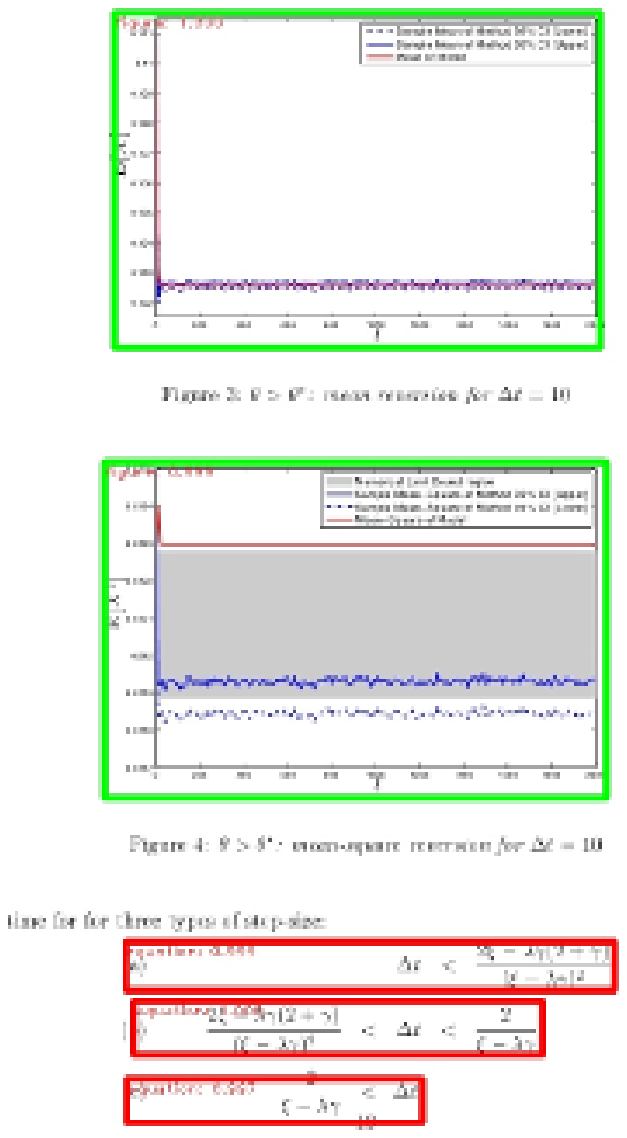, width=0.11\textwidth,height=0.14\textwidth}}
\tcbox[sharp corners, size = tight, boxrule=0.2mm, colframe=black, colback=white]{
\psfig{figure=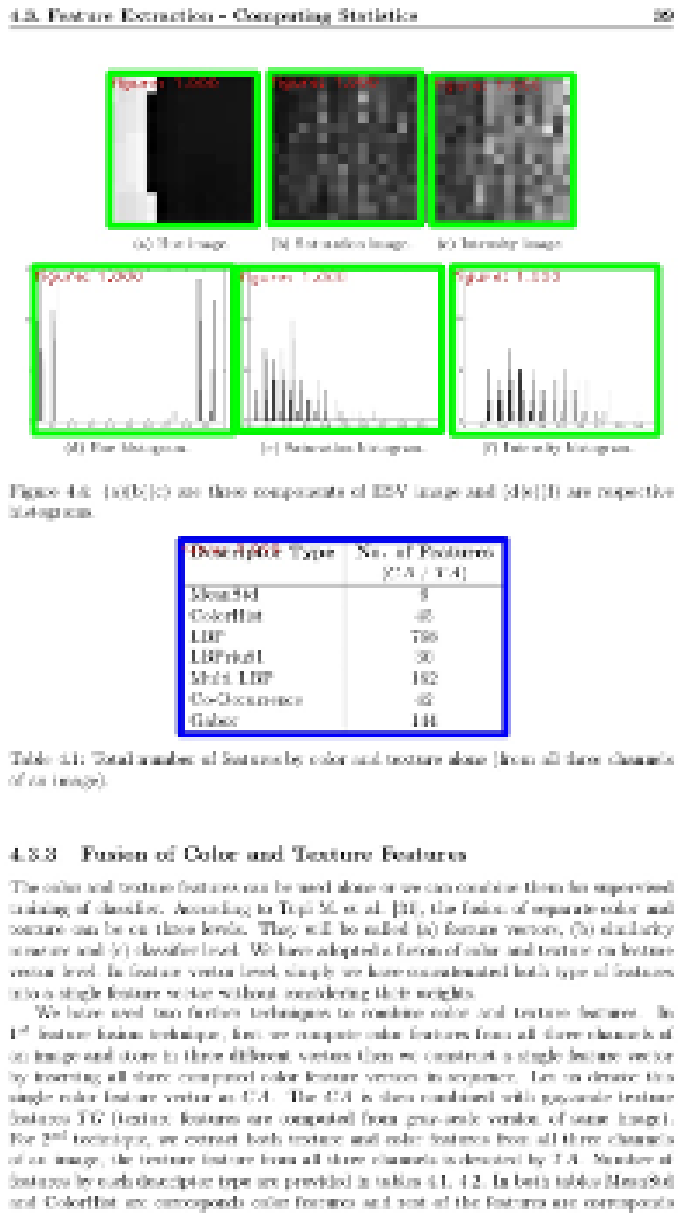, width=0.11\textwidth,height=0.14\textwidth}}}
\caption{Visual illustration of various graphical object detection results of the public benchmark data sets by the \textsc{god} framework. \textbf{First Row:} detection of various forms of tables in various documents. \textbf{Second Row:} detection of multiple graphical objects: tables, equations and figures with various forms and multiple numbers in a single page. Blue, Green and Red colors represent the bounding boxes of table, figure and equation, respectively.\label{fig:result_intro}}
\end{figure}

In this paper, our objective is to detect graphical objects like tables, figures, equations from varied types of digitally generated documents. This problem is conceptually similar to object detection in natural scene images. The diverse layouts and structure of equations, tables and figures make the detection difficult for the rule based systems~\cite{chen_2011_table,tupaj1996extracting,fang2011table,shafait2010table}. Tables have different structures such as cell separated table, table with no separation lines between the rows and columns or alternating colours to separate the cells. All these make the table detection itself very challenging. Sometimes, tables and figures have a high degree of inter-class similarity between themselves and also among other graphical objects, e.g. some charts and plots with several intersections of horizontal and vertical lines will resemble the structure of the tables. Traditional rule based methods have had difficulties to detect them with high precision~\cite{hao2016table,schreiber2017deepdesrt,gilani2017table,kavasidis2018saliency}. 

In this paper, we present a novel end-to-end trainable deep learning based framework, called as Graphical Object Detection (\textsc{god}), for detecting graphical objects particularly tables, figures and equations in the document images. Our presented framework based on the recent object detection algorithms in computer vision~\cite{ren2015faster,redmon2017yolo9000,he2017mask} is data-driven and independent of any heuristic rules for detecting graphical objects in the document images. Usually, deep learning based approaches require a large amount of labelled training data which is not available in our task. To solve the scarcity of labelled training data, the {\sc god} explores the concept of transfer learning and domain adaptation for graphical object detection task in the document images. Experiments on the various public benchmark data sets conclude that the \textsc{god} is superior than state-of-the-art techniques for localizing various graphical objects in the document images. Figure~\ref{fig:result_intro} visually illustrates the success of \textsc{god} technique for localizing various graphical objects present in the document images. In particular, the contributions of this work are as follows:

\begin{itemize}
\item We present an end-to-end trainable deep learning based approach to localize graphical object in the document images inspired by the concept of recent object detection algorithms in computer vision~\cite{ren2015faster,he2017mask}.   

\item We perform transfer learning to fine-tune a pre-trained model for our graphical object detection task in the document images.

\item Our \textsc{god} framework obtains the superior results on public benchmark \textsc{\textbf{icdar-pod2017}} (please see Table~\ref{table_icdar_2017_ap1}), \textsc{\textbf{icdar-2013}} (please see Table~\ref{table_icdar_2013_ap}) and \textsc{\textbf{unlv}} (please see Table~\ref{table_unlv_data}) than state-of-the-arts. 
\end{itemize}

\section{Related Work} \label{related_work}

Automatic information extraction from the digital documents requires the detection and understanding of graphical objects such as tables, figures, equations, etc. A good number of researchers have contributed to localize various page objects and layout analysis on the different types of documents.

\subsection{Rule based Methods}

Before deep learning era, most of the developed approaches~\cite{chen_2011_table, tupaj1996extracting, fang2011table, shafait2010table} for table detection were based on the assumption of table structures and exploiting a prior knowledge on object properties by analyzing tokens extracted from the documents. Tupaj {\em et al.}~\cite{tupaj1996extracting} developed an optical character recognition ({\sc ocr}) based table recognition system which initially identifies the potential table region through the combination of white space and keyword analysis. Keyword analysis determines the header which is regarded as the starting line of the table. The vertical and horizontal passes through the document page identify the white spaces between the cells of the table region. This rule based method is highly dependent on the set of keywords used for analyzing the headers and the inherent assumption about the structure of the tables. Fang {\em et al.}~\cite{fang2011table} analyzed the page layout and detected the table from {\sc pdf} documents based on separator mining which identifies the visual separators of the cells. This method assumes that the table cells are separated either by ruling lines or by white spaces. The disadvantage of such rule based methods is that they heavily rely upon the type and structure of tables under consideration, hence fail to detect the tables whose layout structures are different.

\subsection{Deep Learning based Methods}

Deep Convolutional Neural Networks ({\sc dcnn}s) have proved to be useful in highly complex computer vision tasks to identify visually distinguished features of images. The effectiveness of {\sc dcnn}s are also being explored in recent years in document object analysis by various research groups~\cite{hao2016table,schreiber2017deepdesrt,gilani2017table,kavasidis2018saliency}. Hao {\em et al.}~\cite{hao2016table} applied deep learning for detection of tables in {\sc pdf} documents. They considered heuristic rules to propose regions with table-like structures in the document and then classified them into table or non-table regions using a {\sc cnn}. However, the shortcomings of using rule to identify table-like structures are not yet overcome completely by this method. Schreiber {\em et al.}~\cite{schreiber2017deepdesrt} proposed a deep learning based solution for table detection and structure identification, which does not require any assumption about the structure of the tables. They fine-tuned Faster {\sc r-cnn}~\cite{ren2015faster} with two different backbone architectures: {\sc zfn}et~\cite{zeiler2014visualizing} and {\sc vgg}-16 ~\cite{simonyan2014very} for the table detection task. Gilani {\em et al.}~\cite{gilani2017table} also used the same approach. However, they applied image transformation by stacking three different distance transformed layers before passing them through Faster {\sc r-cnn}. This model can deal with only identifying table regions in the documents.

\begin{figure*}[ht!]
\centerline{
\psfig{figure=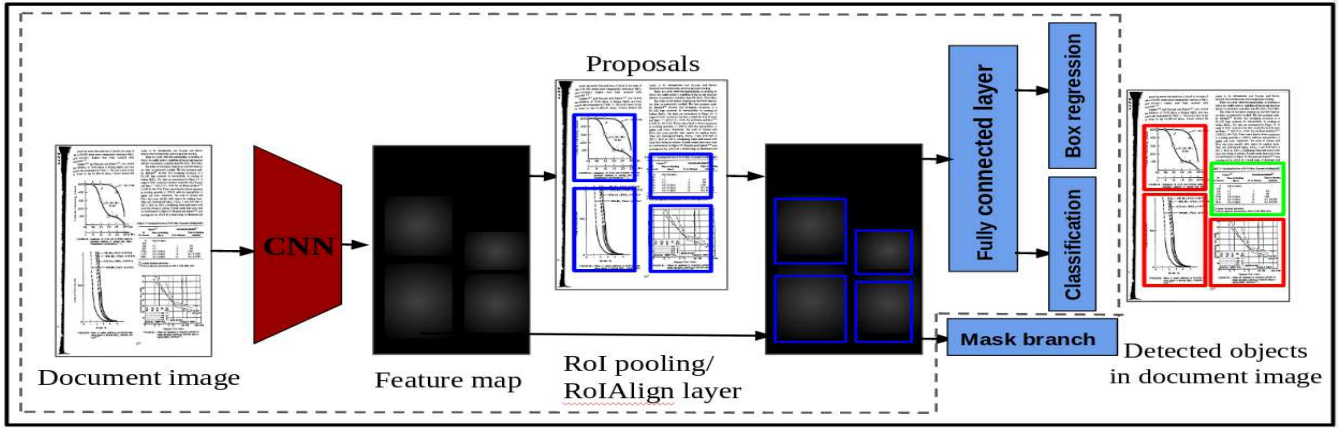,height=0.32\textwidth, width=1.0\textwidth}}
\caption{\textsc{god} framework. The model takes an image as input and generates the feature map. \textsc{rpn} proposes the region based on the feature map. Detection network uses the proposed regions and feature map to detect various graphical objects.\label{fig_diagram}}
\end{figure*}

Several algorithms \cite{okun1999page, moll2008segmentation, nayef2015text, fletcher1988robust, tombre2002text} have been developed to identify text and non-text or graphical regions from document images. As per~\cite{le2015text}, the classification techniques can be broadly classified into three approaches (i) region or block based, (ii) pixel based and (iii) connected component based classification methods. However, none of these methods classify the graphical regions in the document images, which is a complex task as the structures of the graphical regions donot follow any general rule. Kavasidis {\em et al.}~\cite{kavasidis2018saliency} proposed a saliency based {\sc cnn} architecture for document layout analysis which used semantic segmentation to classify each pixel into different classes. This model at first extracted saliency features corresponding to the text, table and figure regions. The extracted segmentation maps by the saliency detector are passed through binary classifiers to classify them. This model used a binary classifier for each of the classes, which are trained separately. This model is able to detect four different kinds of page objects: table, bar chart, pie chart and line chart. Xiaohan Yi {\em et al.}~\cite{yi2017cnn} proposed a dynamic programming based region proposal method for page object detection. 

\section{Graphical Object Detection}

This section presents the details about the proposed \textsc{god} algorithm inspired by the recent object detection algorithms in computer vision~\cite{ren2015faster,he2017mask}. Detection of graphical objects like tables, figures, equations, etc. is basically localization of these objects within a document image. The problem is conceptually similar to the detection of objects in natural scene images. Natural scene objects have visually identifiable characteristics, those features can be extracted by using {\sc cnn}. Similarly graphical objects can also be identified by the specific characteristics of them, which {\sc cnn} is capable of. The compelling performances of Faster {\sc r-cnn} and Mask {\sc r-cnn} on {\sc pascal voc}~\cite{Everingham10} and \textsc{coco}~\cite{lin2014microsoft} data sets makes us adapting the framework into our graphical object detection work. Training of deep networks requires a large amount of data, which we lack in the document object detection domain. Hence, we consider domain adaptation and transfer learning in our work. We test the abilities of Faster {\sc r-cnn} and Mask {\sc r-cnn}, originally built for the natural scene images, to cope with detecting graphical objects in the document images. The performances of ImageNet~\cite{imagenet_cvpr09} pre-trained models as the backbone of \textsc{god} model inspired by faster {\sc r-cnn} is listed in Table~\ref{table_resnet_model}.

Figure~\ref{fig_diagram} displays the overview of the \textsc{god} method. Dotted region in Figure~\ref{fig_diagram} is the Faster {\sc r-cnn} block. As described by \cite{ren2015faster}, Faster-{\sc rcnn} is comprised of two modules. The first module is Region Proposal Network ({\sc rpn}), responsible for proposing regions which might contain objects. The second module, Fast-{\sc rcnn} detector~\cite{girshick2015fast}, detects the objects using the proposed regions. {\sc rpn} attracts the attention of the Fast-{\sc rcnn} as to look into the proposed regions. The whole system works a single, unified network for the object detection task. Similar to Faster {\sc r-cnn}, Mask {\sc r-cnn}~\cite{he2017mask} adopts the same two stage procedure (i) Region Proposal Network ({\sc rpn}) for proposing regions and (ii) prediction of the class label and bounding box regression along with binary mask of each region of interests.

Region Proposal Network ({\sc rpn}) proposes rectangular regions, each associated with objectness score~\cite{ren2015faster} which tells the detector whether the region contains an object or the background. This module is capable of handling input images of any size. {\sc rpn} module is also \textit{translation-invariant}. Translation invariant means if an object in the image is translated, the proposal should also be translated and the same {\sc rpn} module should predict the region proposal in either of the location.

The input image is first passed through a convolutional layer which generates a feature map. The Region of Interest (RoI) pooling layer in case of Faster {\sc r-cnn} performs max pooling to convert the nonuniform size inputs to fixed-size feature maps. The output feature vector is then passed through fully connected layers: box-regression (\textit{reg}) and classification (\textit{cls}) layer for predicting object bounding box and class label. While in case of Mask {\sc r-cnn}, RoIAlign layer generates fixed sized feature by preserving the exact spatial locations. Finally, the fixed sized features pass to two different modules: multiple layer perceptron to predict object bounding box and class label and mask module to predict segmentation mask.

\subsection{Implementation Details}

The model is trained and tested on various data sets with fixed image sizes, but the dimensions of the images from a particular data set vary in a range. The images are then resized to a fixed size of $600 \times 600$ before passing through the Faster-{\sc rcnn}.
The model is implemented using PyTorch and trained and tested on Nvidia GeForce {\sc gtx} $1080$ Ti {\sc gpu}s with batch size of $4$. We used Caffe pre-trained models of {\sc vgg}-16~\cite{simonyan2014very} and ResNets~\cite{he2016deep}, trained on ImageNet~\cite{imagenet_cvpr09}, as the backbone of the Faster {\sc r-cnn}. For {\sc vgg}-16, the last max pooling layer is not used in the model.

Since, the effective receptive field of the input image is large, we use $3 \times 3$ sliding windows in the {\sc rpn}. To generate $k$=$30$ anchor boxes, we considered $6$ different anchor scales in powers of $2$ from $8$ to $512$ and anchor ratios of $1$ through $5$ so that the region proposals can cover almost every part of the image irrespective of the image size. We used stochastic gradient descent {\sc sgd} as an optimizer with initial learning rate $= 0.001$ and the learning rate decays after every $5$ epochs and it is equal to $0.1$ times of the previous value. For further implementation and architecture details, please refer to the source code at: \url{https://github.com/rnjtsh/graphical-object-detector}. 

\section{Experiments}

\subsection{Evaluation Measures}

We use mean average precision (m{\sc ap})~\cite{Everingham10}, average precision, recall and $F1$ measures~\cite{tran2015table,schreiber2017deepdesrt,kavasidis2018saliency} to evaluate the performance of our algorithm for detecting graphical objects in the document images. 

\subsection{Public Data Sets}

\paragraph{\sc \textbf{icdar-pod2017}~\cite{gao2017icdar2017}}

This data set consists of $2417$ English document images selected from $1500$ scientific papers of CiteSeer. It includes large variety in both layout and object styles: single-column, double-column, multi-column and various kinds of equations, tables and figures. This data set is divided into training set consisting of $1600$ and test set consisting of $817$ images.   

\paragraph{\sc \textbf{icdar-2013}~\cite{gobel2013icdar}}

This data set contains $67$ {\sc pdf}s with $150$ tables - $40$ {\sc pdf}s excerpt from the {\sc us} Government and $27$ {\sc pdf}s from {\sc eu}. We use converted images corresponding to {\sc pdf}s for our experiment. We use this data set only for testing purpose due to limited to number of images.

\paragraph{\textbf{Marmot table recognition data set}}\footnote{\url{http://www.icst.pku.edu.cn/cpdp/sjzy/index.htm}}

It consists of $2000$ document images with a great variety in page layout and table styles. We use this data set only for training our network.

\paragraph{{\sc \textbf{unlv}} \textbf{data set}~\cite{shahab2010open}}

It contains $2889$ pages of scanned document images from variety of sources (Magazines, Newspapers, Business letters, Annual reports, etc). Among them, only $427$ images contain table zone. We consider $427$ images for evaluation purpose.

\subsection{Ablation Study}

We conduct a number of ablation experiments in the context of document object detection to quantify the importance of each of the components of our algorithm and to justify various design choices. Our {\sc god} (Faster {\sc r-cnn}) uses {\sc \textbf{icdar-pod2017}} data set for this purpose. 

\paragraph{\textbf{Pre-trained Model}}

Researchers have already proven that deeper neural networks are beneficial for large scale image classification. To further analyze the \textsc{god} (Faster {\sc r-cnn}), we conduct an experiment with different depths of pre-trained models. The details of detection scores of the \textsc{god} (Faster {\sc r-cnn}) with different pre-trained models with same settings are listed in Table~\ref{table_resnet_model}. The table quantitatively shows that with the same setting, the deeper network obtains better detection accuracy.      
\begin{table}[ht!]
\begin{center}
\begin{tabular}{|l |l l l |l |} \hline
\textbf{Models} & \multicolumn{3}{ l| }{\textbf{Test Performance: AP}}  &  \textbf{mAP} \\\cline{2-4} 
  & \textbf{Equation} & \textbf{Table} & \textbf{Figure} & \\ \hline
VGG-16      &0.807 &0.934 &0.857 &0.866  \\
ResNet-50   &0.817 &0.950 &0.819 &0.862  \\ 
ResNet-101  &0.894 &0.959 &0.857 &0.899 \\
ResNet-152  &\textbf{0.903} &\textbf{0.962}	&\textbf{0.851} &\textbf{0.905} \\ \hline \end{tabular}
\end{center}
\caption{Performances of different backbone models on {\sc \textbf{ icdar-pod2017}} with IoU threshold $0.6$. Deeper pre-trained model obtains higher detection accuracy. \textsc{\textbf{ap}}: average precision. Bold value indicates the best result.\label{table_resnet_model}} 
\end{table}

\paragraph{\textbf{IoU Threshold}}

We conduct a study on the detection performance of the \textsc{god} (Faster {\sc r-cnn}) with varying IoU threshold. Table~\ref{table_icdar_2017_iou} shows the detection statistics of the \textsc{god} (Faster {\sc r-cnn}) on varying IoU threshold. With the same setting, increasing threshold value can reduce the detection accuracy (i.e. values of m{\sc ap} and Ave F1). This is because of reduction in the number of true positive when increasing IoU threshold. From the Table~\ref{table_icdar_2017_iou}, we observed that we obtained the best m{\sc ap} and Ave F1 when the IoU threshold is set to $0.5$.    
\begin{table}[ht!]
\addtolength{\tabcolsep}{-2.5pt}
\begin{center}
\begin{tabular}{|l| l l l |l |l l l |l|} \hline
\textbf{IoU} & \multicolumn{3}{l|}{\textbf{Test Performance: AP}}  &  \textbf{mAP} & \multicolumn{3}{l|}{\textbf{Test Performance: F1}}  &  \textbf{Ave F1}\\\cline{2-4} \cline{6-8}
  & \textbf{Eqn.} & \textbf{Table} & \textbf{Figure} &  &\textbf{Eqn.} & \textbf{Table} & \textbf{Figure} &  \\\hline  
0.5 &0.922 &0.980 &0.881 &0.928  &0.922 &0.977	&0.851	 &0.917 \\
0.6 &0.903 &0.962 &0.851 &0.905  &0.889  &0.959	&0.878	&0.909  \\
0.7 &0.810 &0.937 &0.815 &0.854 &0.832 &0.952 &0.833 &0.872   \\
0.8 &0.742 &0.916  &0.786 &0.814 &0.772 &0.927 &0.833 &0.844   \\ \hline
\end{tabular}
\end{center}
\caption{Greater IoU threshold reduces the performance of the \textsc{god} (Faster {\sc r-cnn}) model on {\sc \textbf{ icdar-pod2017}} due to reduction in number of true positives.\label{table_icdar_2017_iou}} 
\end{table}

\begin{table*}[ht!]
\addtolength{\tabcolsep}{-4.0pt}
\begin{center}
\begin{tabular}{|l| l l l |l |l l l |l| l l l |l |l l l |l|} \hline
\multicolumn{9}{|c|}{IoU$=0.8$} & \multicolumn{8}{c|}{IoU$=0.6$}\\ \hline
\textbf{Methods} & \multicolumn{3}{l|}{\textbf{Test Performance: AP}}  &  \textbf{mAP} & \multicolumn{3}{l|}{\textbf{Test Performance: F1}}  &  \textbf{Ave F1} & \multicolumn{3}{l|}{\textbf{Test Performance: AP}}  &  \textbf{mAP} & \multicolumn{3}{l|}{\textbf{Test Performance: F1}}  &  \textbf{Ave F1} \\\cline{2-4} \cline{6-8} \cline {10-12} \cline{14-16}
  & \textbf{Eqn.} & \textbf{Table} & \textbf{Figure} & &\textbf{Eqn.} & \textbf{Table} & \textbf{Figure}&  & \textbf{Eqn.} & \textbf{Table} & \textbf{Figure} &  &\textbf{Eqn.} & \textbf{Table} & \textbf{Figure} & \\\hline  
{\sc NLPR-PAL}~\cite{gao2017icdar2017}	&0.816 	&0.911 	 &0.805 &0.844 &0.902 &0.951 &0.898 &0.917 &0.839 &0.933	&0.849 &0.874 &0.915 &0.960 &0.927 &0.934 \\
HustVision~\cite{gao2017icdar2017}	&0.293 	&0.796 	 &0.656  &0.582 &0.042 	&0.115 	&0.132 	&0.096 &0.854 	&0.938 	&0.853 	&0.882 &0.078 	&0.132 	&0.164 	&0.124 \\
FastDetectors~\cite{gao2017icdar2017} &0.427 & 0.884 &0.365  &0.559 &0.639 	&0.896 	&0.616 	&0.717  &0.474 	&0.925 	&0.392 	&0.597 &0.675 	&0.921 	&0.638 	&0.745 \\
Vislnt~\cite{gao2017icdar2017} &0.117 & 0.795 &0.565  &0.492 &0.241 	&0.826 	&0.643 	&0.570 &0.524 	&0.914 	&0.781 	&0.740 &0.605 &0.921 &0.824 &0.783 \\
\textsc{god} (Faster {\sc R-CNN}) & 0.742 &0.916 & 0.786 &	0.814 & 0.772 & 0.927 & 0.833 &	0.844 &0.903 &0.962 &0.851 &0.905 & 0.889 & 0.959 & 0.878 & 0.909 \\ 
\textsc{god} (Mask {\sc R-CNN}) &\textbf{0.869} &\textbf{0.974} &\textbf{0.818} &\textbf{0.887} &\textbf{0.919} &\textbf{0.968} &\textbf{0.912} &\textbf{0.933} &\textbf{0.921}  &\textbf{0.989} &\textbf{0.860} &\textbf{0.921} &\textbf{0.924} &\textbf{0.971} &\textbf{0.918} &\textbf{0.938} \\ \hline  
\end{tabular}
\end{center}
\caption{Comparison with state-of-the-arts based on m{\sc ap} and Ave F1 with IoU $= 0.8$ and $0.6$, respectively on \textsc{\textbf{icdar-pod2017}} data set. Our {\sc god} (Mask {\sc r-cnn}) is better than state-of-the-arts while IoU $= 0.8$ and IoU $= 0.6$. Bold value indicates the best result.\label{table_icdar_2017_ap1}} \end{table*}

\begin{figure*}[ht!]
\centerline{
\tcbox[sharp corners, size = tight, boxrule=0.2mm, colframe=black, colback=white]{
\psfig{figure=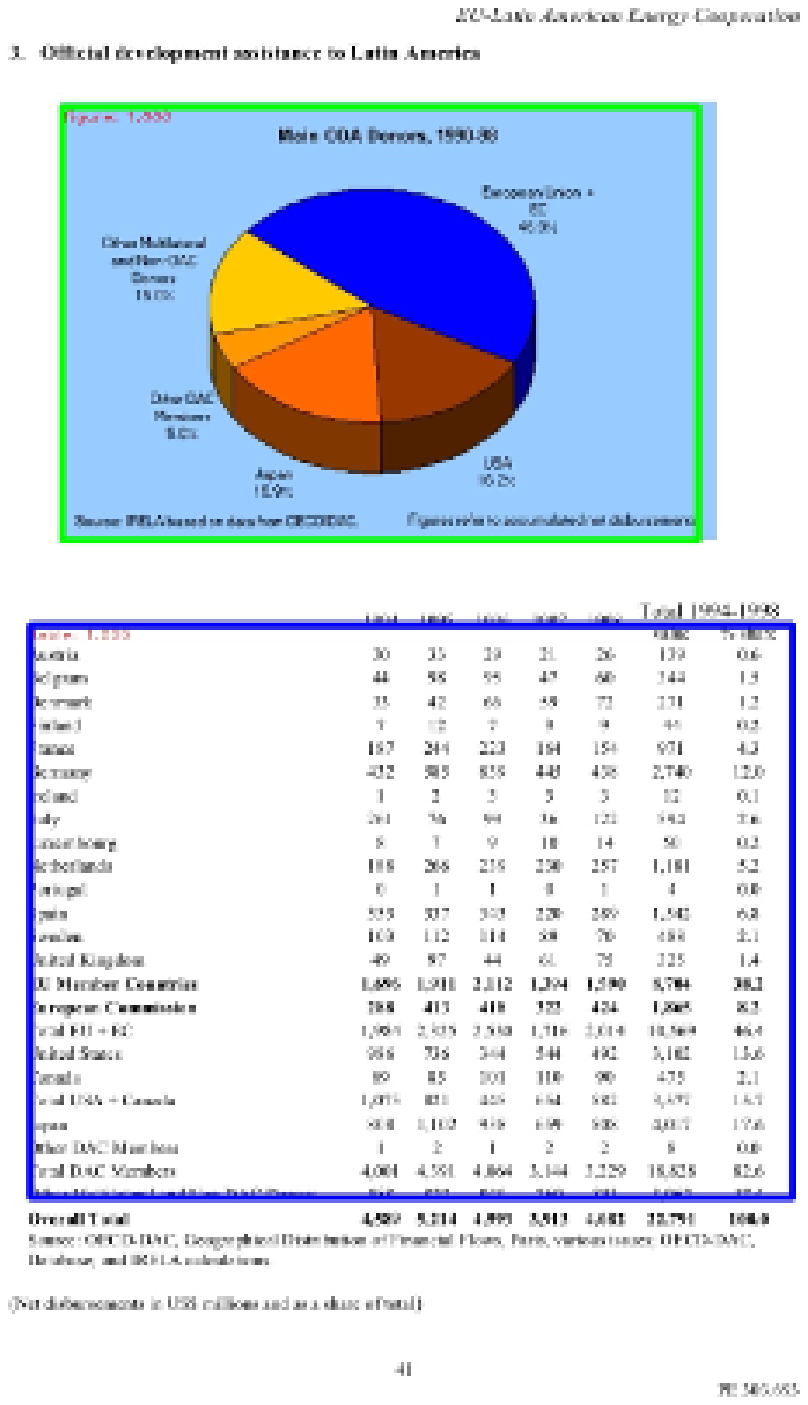, width=0.24\textwidth,height=0.21\textwidth}}
\hspace{-0.01\textwidth}
\tcbox[sharp corners, size = tight, boxrule=0.2mm, colframe=black, colback=white]{
\psfig{figure=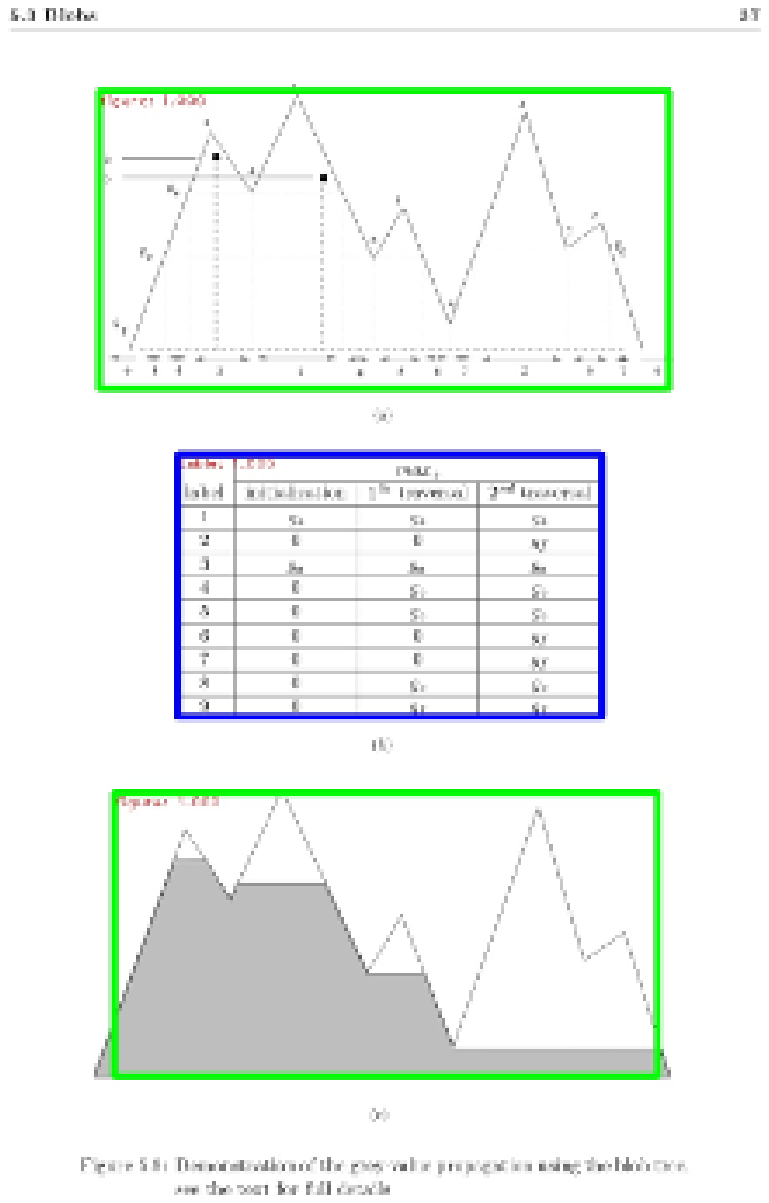, width=0.24\textwidth,height=0.21\textwidth}}
\hspace{0.01\textwidth}
\tcbox[sharp corners, size = tight, boxrule=0.2mm, colframe=black, colback=white]{
\psfig{figure=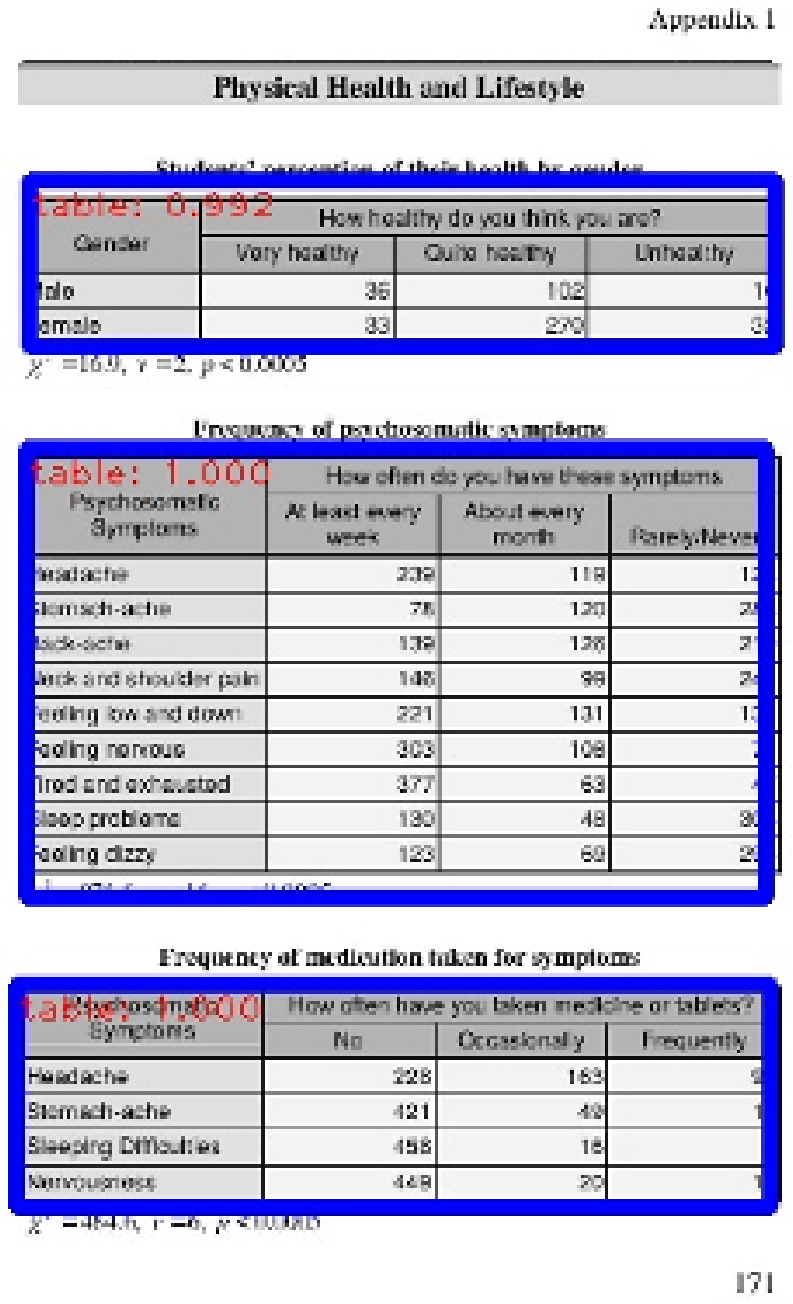, width=0.24\textwidth,height=0.21\textwidth}}
\hspace{0.001\textwidth}
\tcbox[sharp corners, size = tight, boxrule=0.2mm, colframe=black, colback=white]{
\psfig{figure=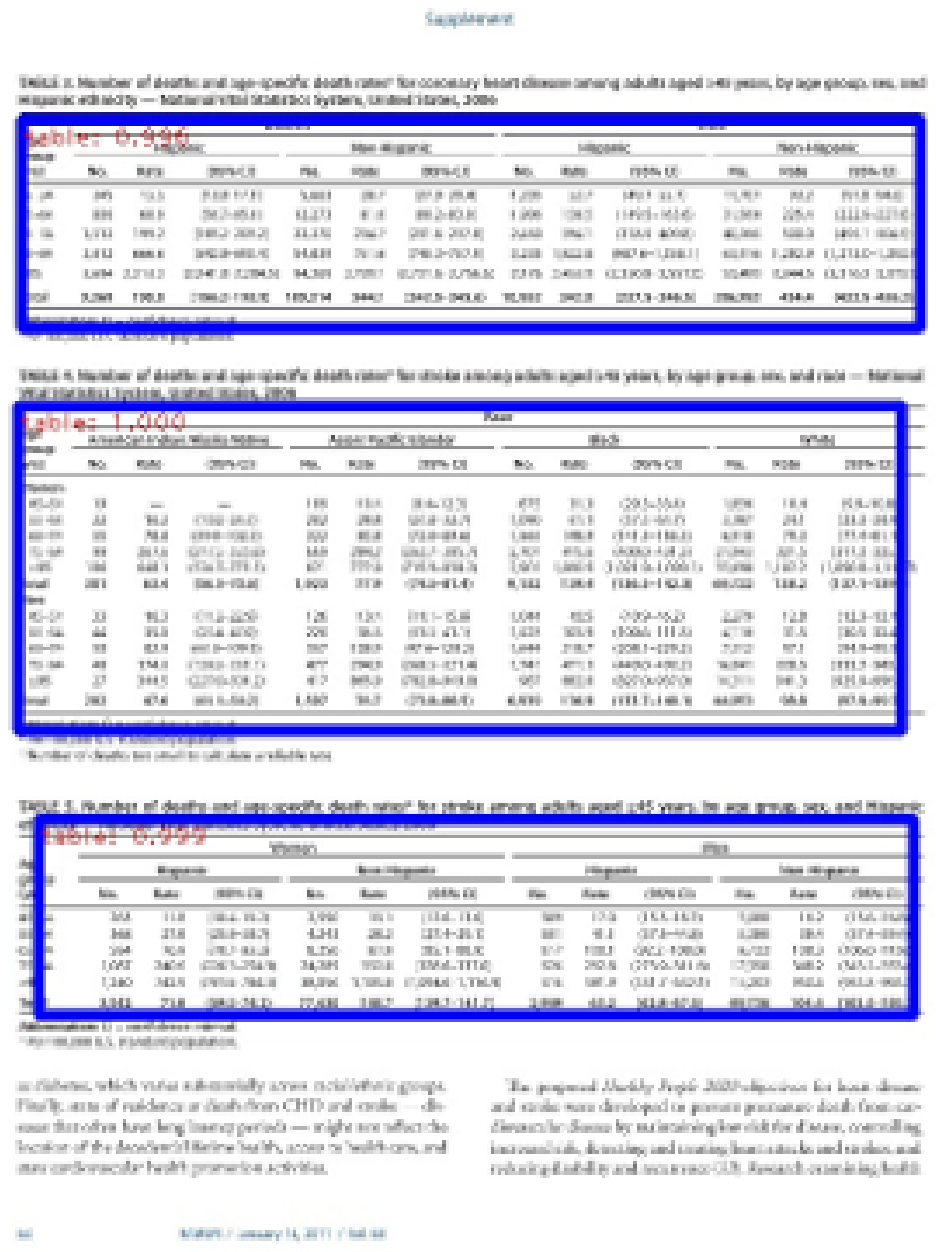, width=0.24\textwidth,height=0.21\textwidth}}}
\vspace{-0.005\textwidth}
\centerline{
\tcbox[sharp corners, size = tight, boxrule=0.2mm, colframe=black, colback=white]{
\psfig{figure=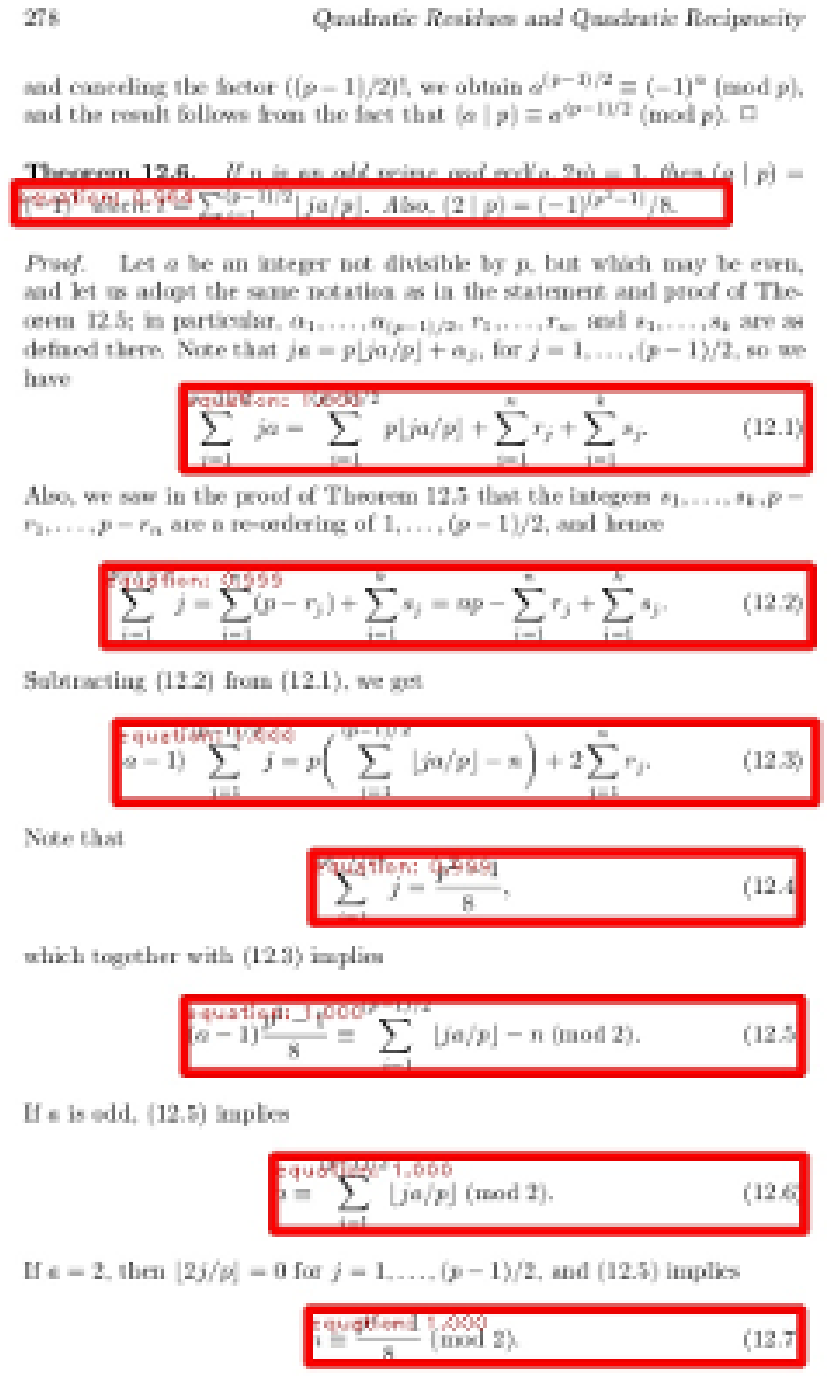, width=0.24\textwidth,height=0.21\textwidth}}
\hspace{-0.01\textwidth}
\tcbox[sharp corners, size = tight, boxrule=0.2mm, colframe=black, colback=white]{
\psfig{figure=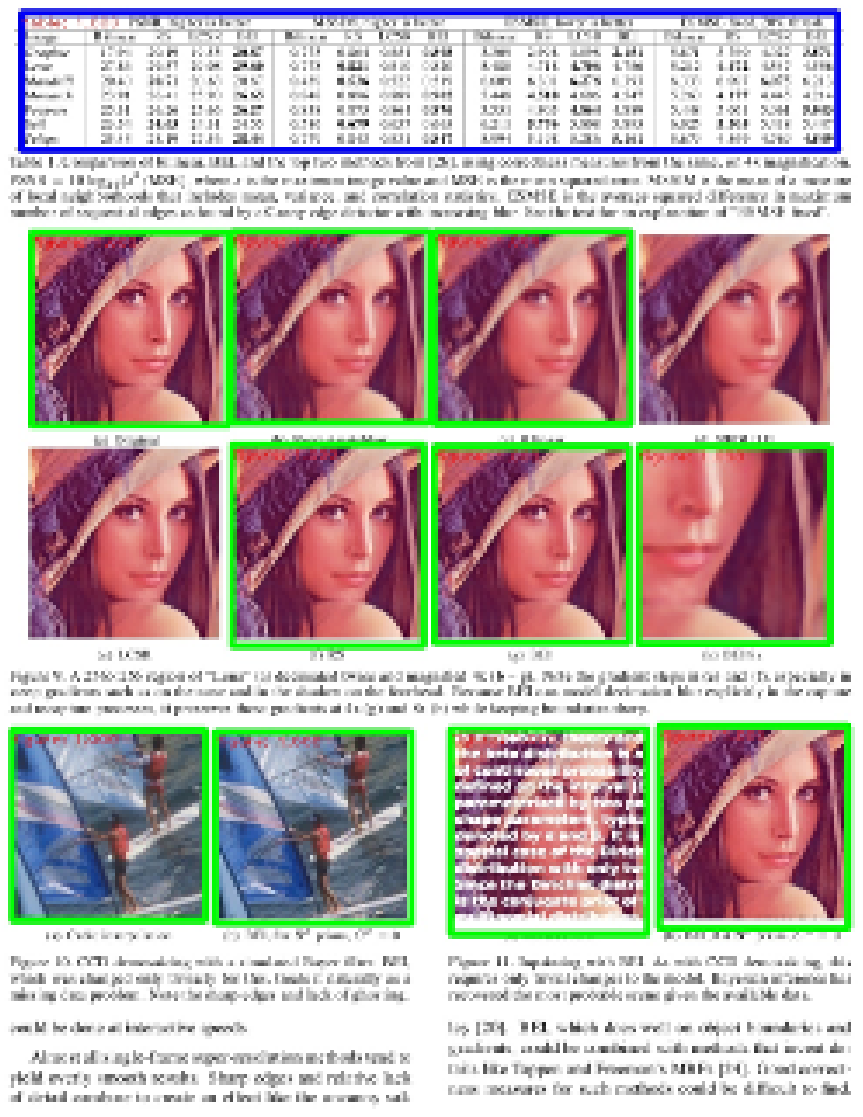, width=0.24\textwidth,height=0.21\textwidth}}
\hspace{0.01\textwidth}
\tcbox[sharp corners, size = tight, boxrule=0.2mm, colframe=black, colback=white]{
\psfig{figure=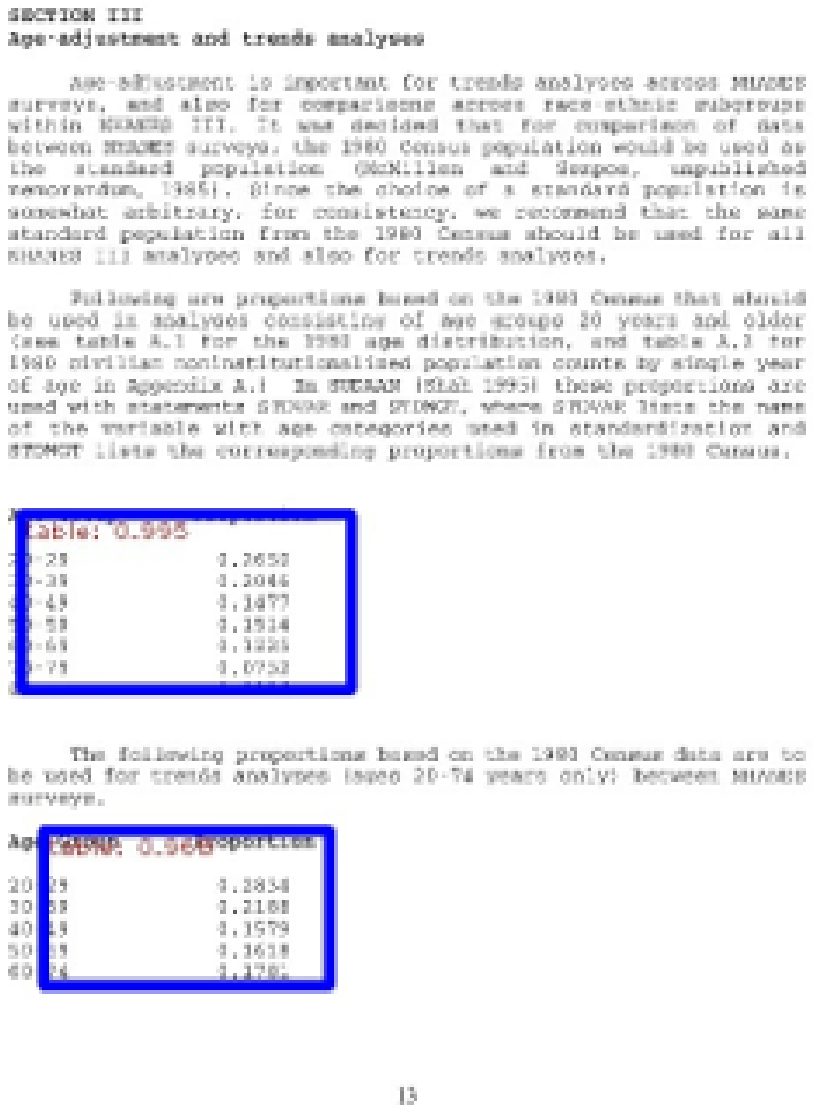, width=0.24\textwidth,height=0.21\textwidth}}
\hspace{-0.01\textwidth}
\tcbox[sharp corners, size = tight, boxrule=0.2mm, colframe=black, colback=white]{
\psfig{figure=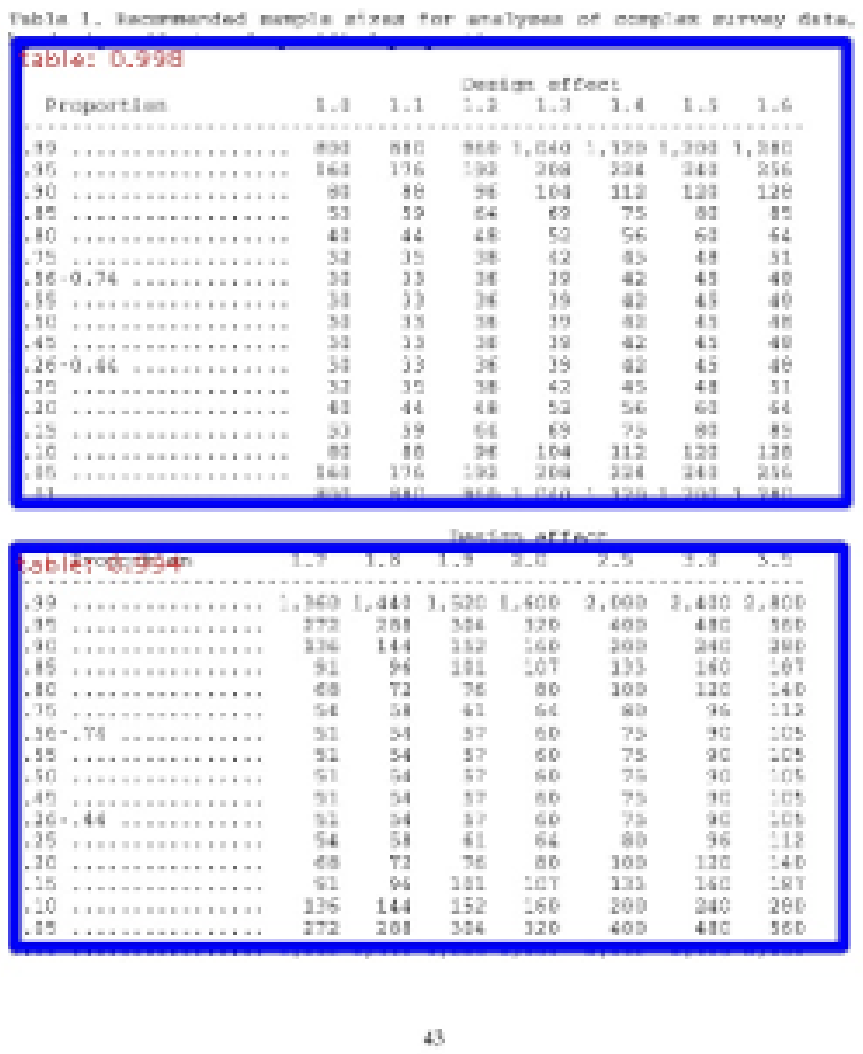, width=0.24\textwidth,height=0.21\textwidth}}}
\vspace{-0.005\textwidth}
\centerline{
\tcbox[sharp corners, size = tight, boxrule=0.2mm, colframe=black, colback=white]{
\psfig{figure=new_icdar2017/POD_1607.eps, width=0.24\textwidth,height=0.21\textwidth}}
\hspace{-0.01\textwidth}
\tcbox[sharp corners, size = tight, boxrule=0.2mm, colframe=black, colback=white]{
\psfig{figure=new_icdar2017/POD_1612.eps, width=0.24\textwidth,height=0.21\textwidth}}
\hspace{0.01\textwidth}
\tcbox[sharp corners, size = tight, boxrule=0.2mm, colframe=black, colback=white]{
\psfig{figure=new_icdar2013/eu-001_1.eps, width=0.24\textwidth,height=0.21\textwidth}}
\hspace{-0.01\textwidth}
\tcbox[sharp corners, size = tight, boxrule=0.2mm, colframe=black, colback=white]{
\psfig{figure=new_icdar2013/us-002_3.eps, width=0.24\textwidth,height=0.21\textwidth}}}
\centerline{(a) \hspace{0.48\textwidth} (b)}
\caption{(a) Results of graphical objects: table, figure and equation localization using the \textsc{god} (Mask {\sc r-cnn}) on {\sc \textbf{icdar-pod2017}} data set. Blue, Green and Red colors represent the predicted bounding boxes of table, figure and equation, respectively. (b) Results of table localization using the \textsc{god} (Mask {\sc r-cnn}) in the document images of {\sc \textbf{icdar-2013}} data set. Blue color represents the predicted bounding box of the table.\label{fig:result_icdar2017}}
\end{figure*}

\subsection{Graphical Object Detection in Document Images}

\paragraph{\textbf{Comparison with the state-of-the-arts on }{\sc \textbf{icdar-pod2017}}}

We compare the \textsc{god} with state-of-the-art techniques on {\sc \textbf{icdar-pod2017}} data set. Methods like {\sc nlpr-pal}, HustVision, FastDetectors and Vislnt submitted to {\sc icdar 2017 pod} competition~\cite{gao2017icdar2017} are considered as state-of-the-art techniques. Table~\ref{table_icdar_2017_ap1} highlights that our method {\sc god} (Mask {\sc r-cnn}) is better than the state-of-the-art approaches with respect to both these measures: m{\sc ap} and Ave F1 when IoU threshold is set to $0.8$ and $0.6$. From the Table, it is observed that the performance of the existing techniques except \textsc{nlpr-pal} are drastically reduced by increasing IoU threshold value. On the other hand, the performance of the \textsc{god} (Mask {\sc r-cnn}) and \textsc{god} (Faster {\sc r-cnn}) are reasonable stable while changing IoU threshold value from $0.6$ to $0.8$. This observation conclude that the \textsc{god} is robust with respect to IoU threshold value.

Figure~\ref{fig:result_icdar2017}(a) displays the results of localization of various graphical objects in the document images of {\sc \textbf{icdar-pod2017}} data set using the \textsc{god} (Mask {\sc r-cnn}). From Figure, it can be observed that the \textsc{god} (Mask {\sc r-cnn}) is able to detect all graphical objects: table, equation and figure with large variability present in a single page. It is also observed that it is able to detect multiple tables, figures and equations in a single page.    

\subsection{Table Detection in Document Images}

\begin{figure*}[ht!]
\centerline{
\tcbox[sharp corners, size = tight, boxrule=0.2mm, colframe=black, colback=white]{
\psfig{figure=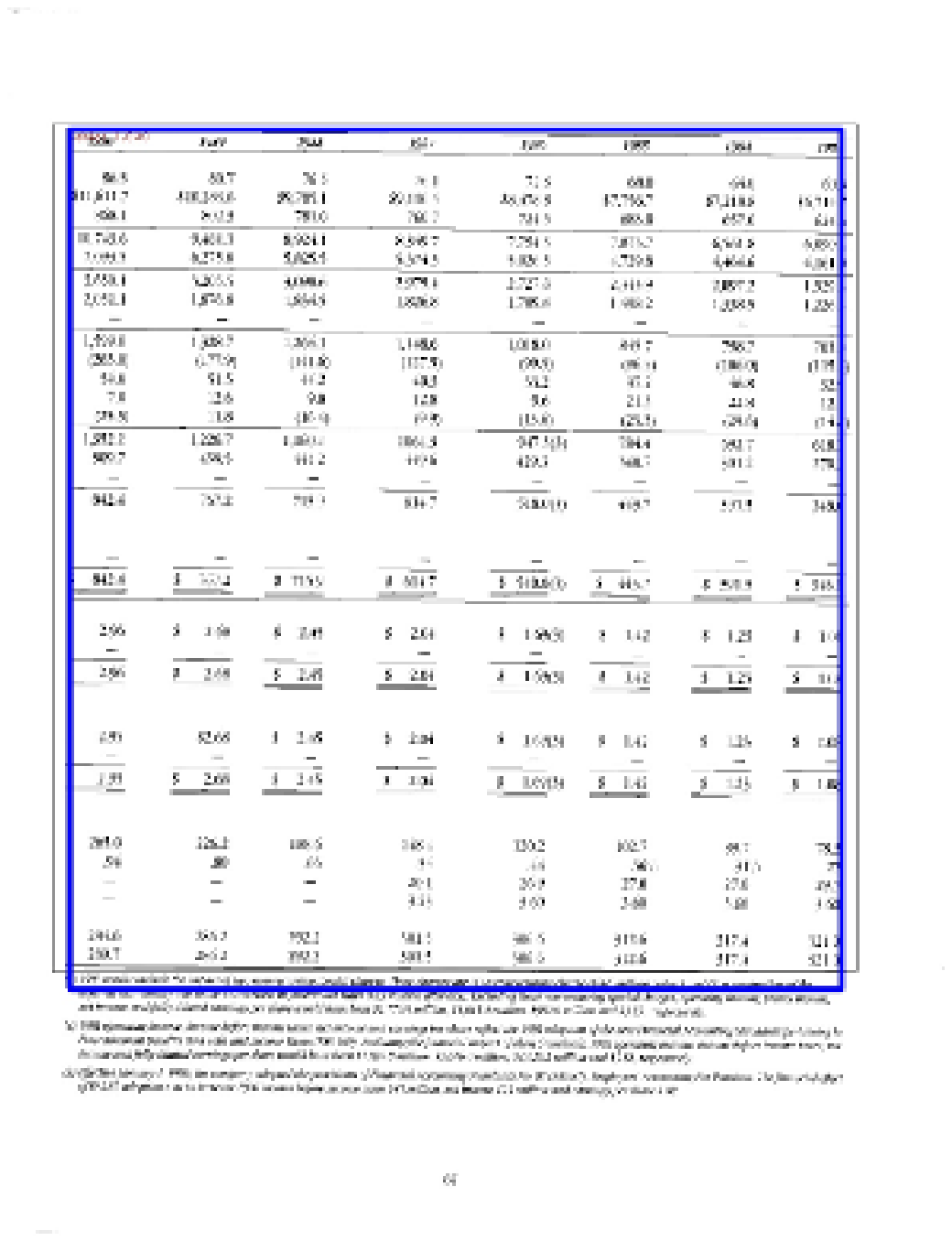, width=0.24\textwidth,height=0.18\textwidth}}
\hspace{-0.01\textwidth}
\tcbox[sharp corners, size = tight, boxrule=0.2mm, colframe=black, colback=white]{
\psfig{figure=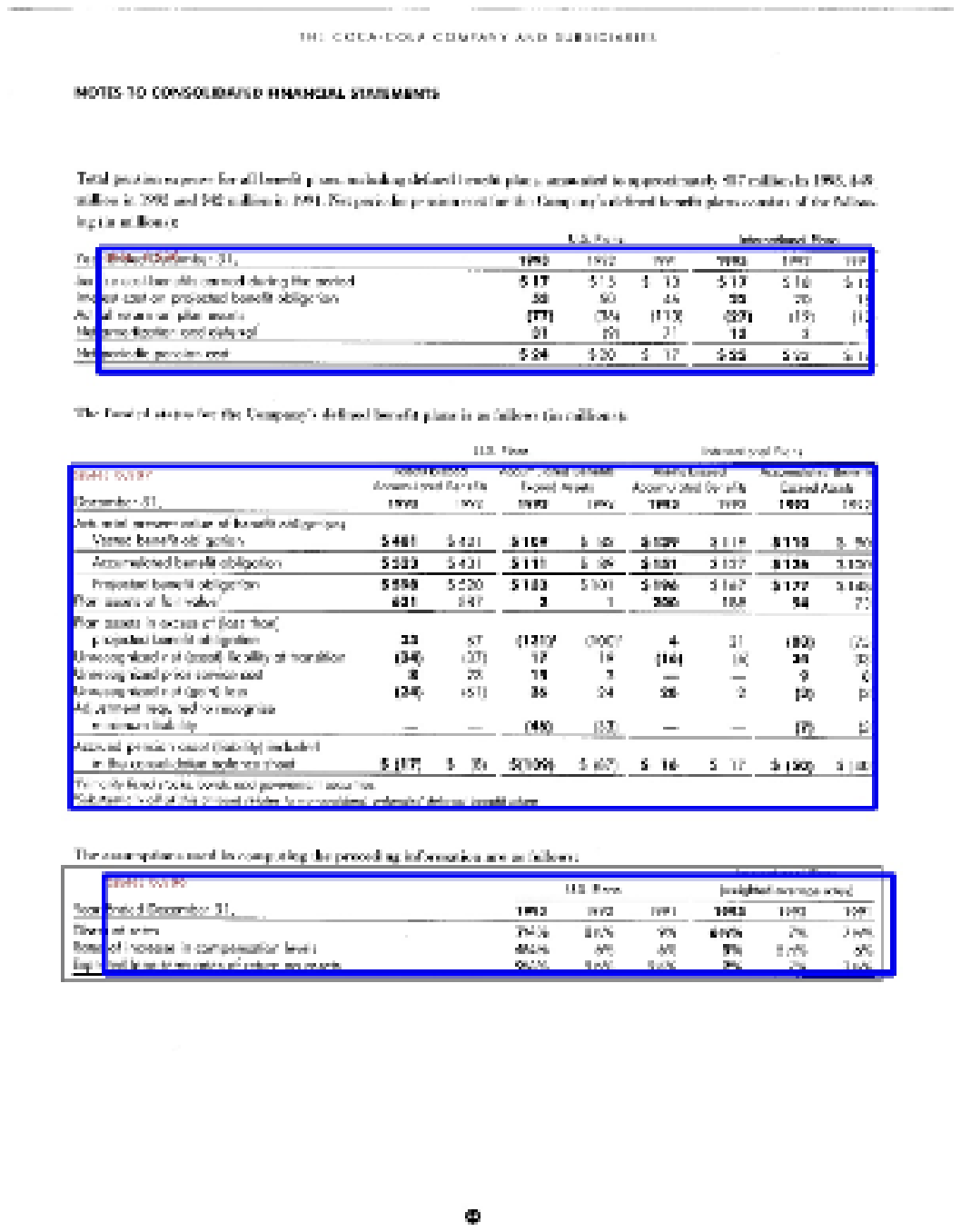, width=0.24\textwidth,height=0.18\textwidth}}
\hspace{0.01\textwidth}
\tcbox[sharp corners, size = tight, boxrule=0.2mm, colframe=black, colback=white]{
\psfig{figure=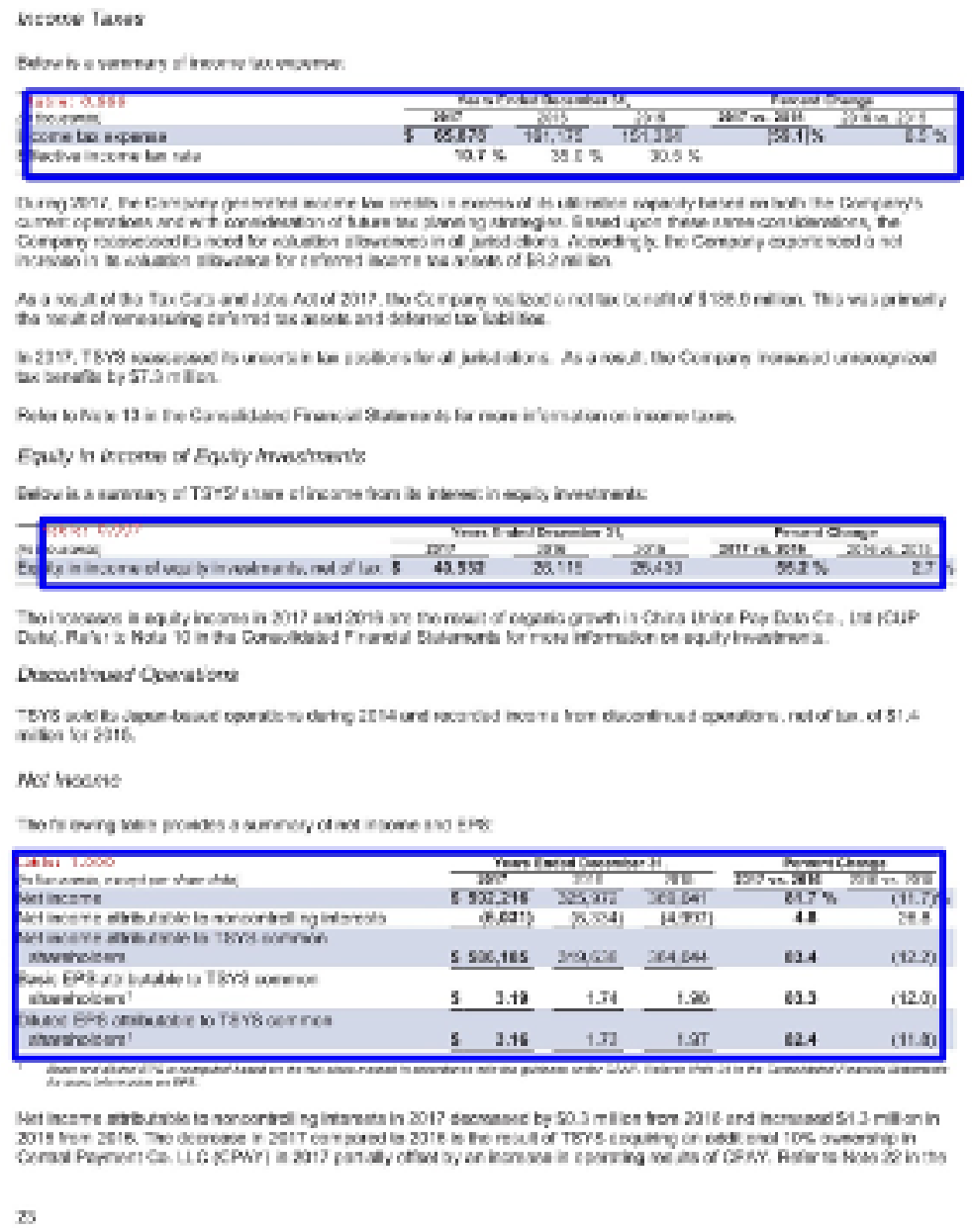, width=0.24\textwidth,height=0.18\textwidth}}
\hspace{-0.01\textwidth}
\tcbox[sharp corners, size = tight, boxrule=0.2mm, colframe=black, colback=white]{
\psfig{figure=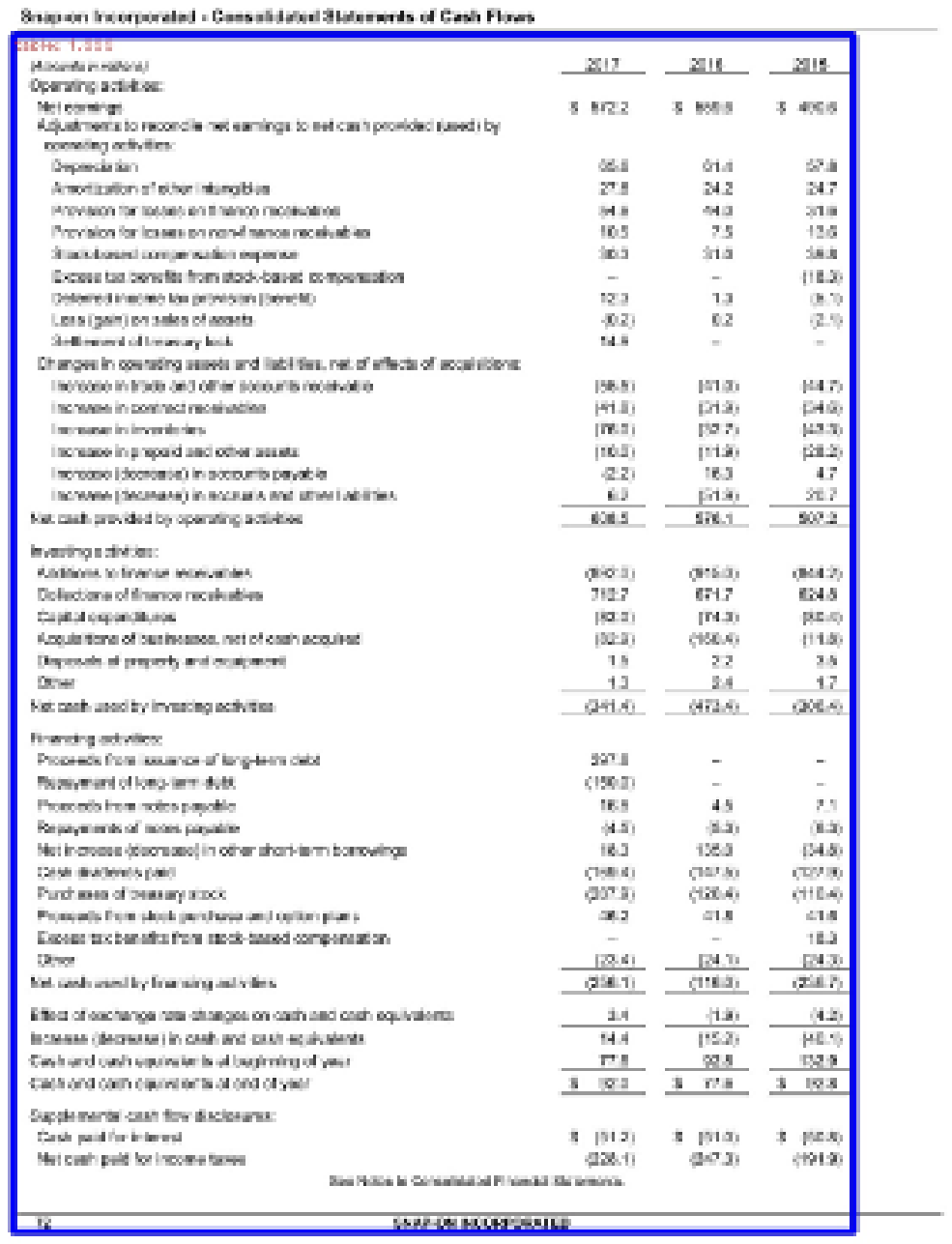, width=0.24\textwidth,height=0.18\textwidth}}}
\vspace{-0.005\textwidth}
\centerline{
\tcbox[sharp corners, size = tight, boxrule=0.2mm, colframe=black, colback=white]{
\psfig{figure=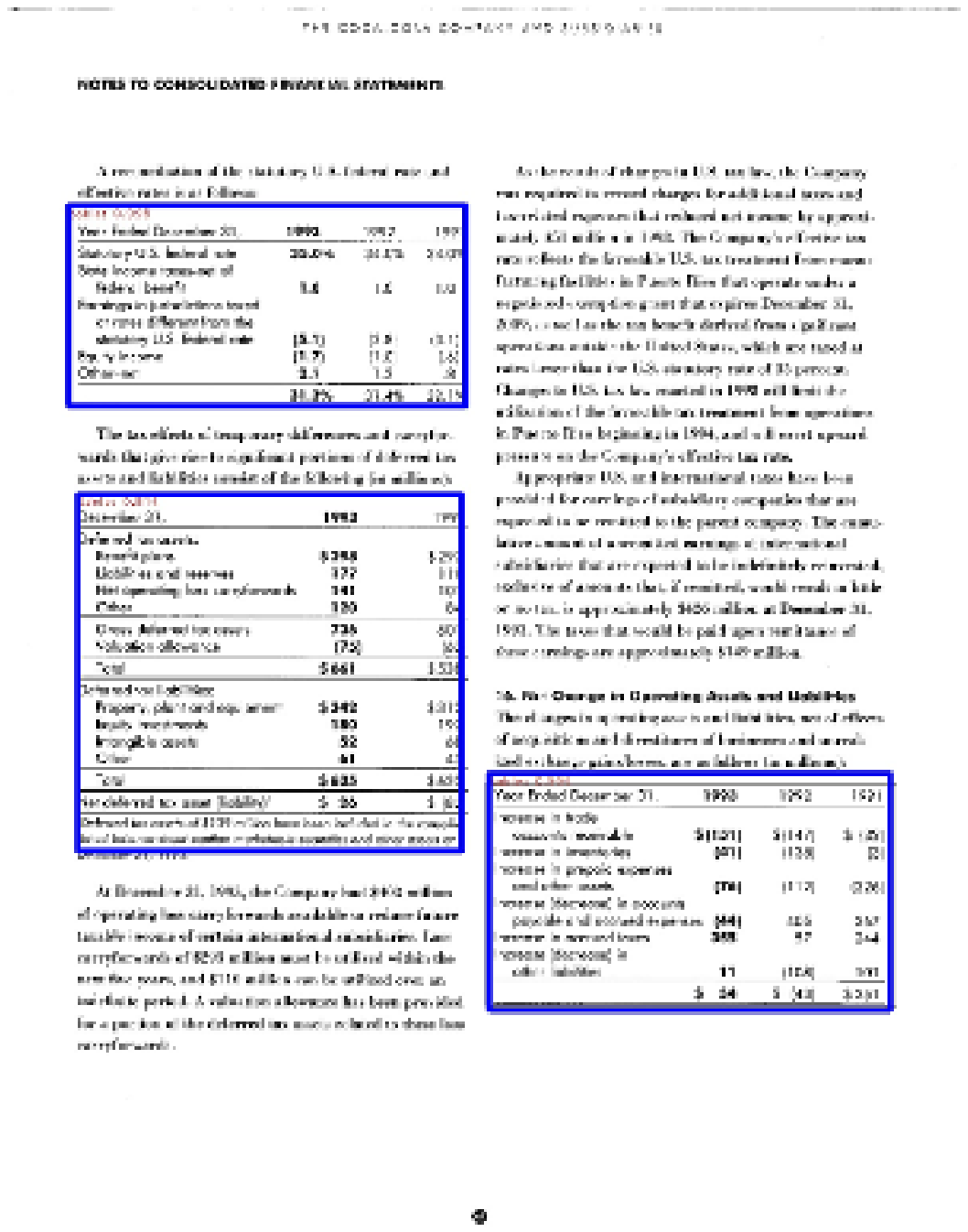, width=0.24\textwidth,height=0.18\textwidth}}
\hspace{-0.01\textwidth}
\tcbox[sharp corners, size = tight, boxrule=0.2mm, colframe=black, colback=white]{
\psfig{figure=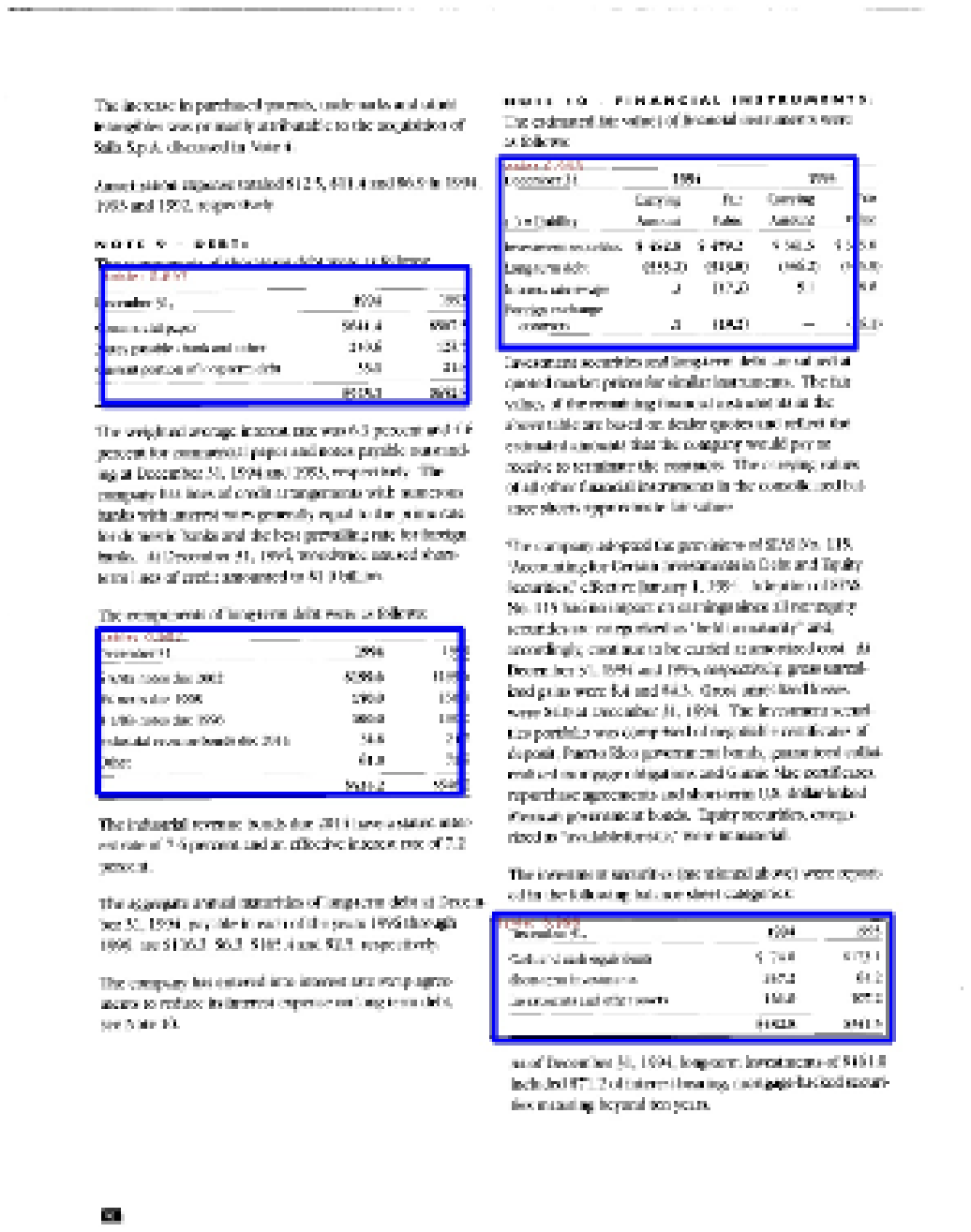, width=0.24\textwidth,height=0.18\textwidth}}
\hspace{0.01\textwidth}
\tcbox[sharp corners, size = tight, boxrule=0.2mm, colframe=black, colback=white]{
\psfig{figure=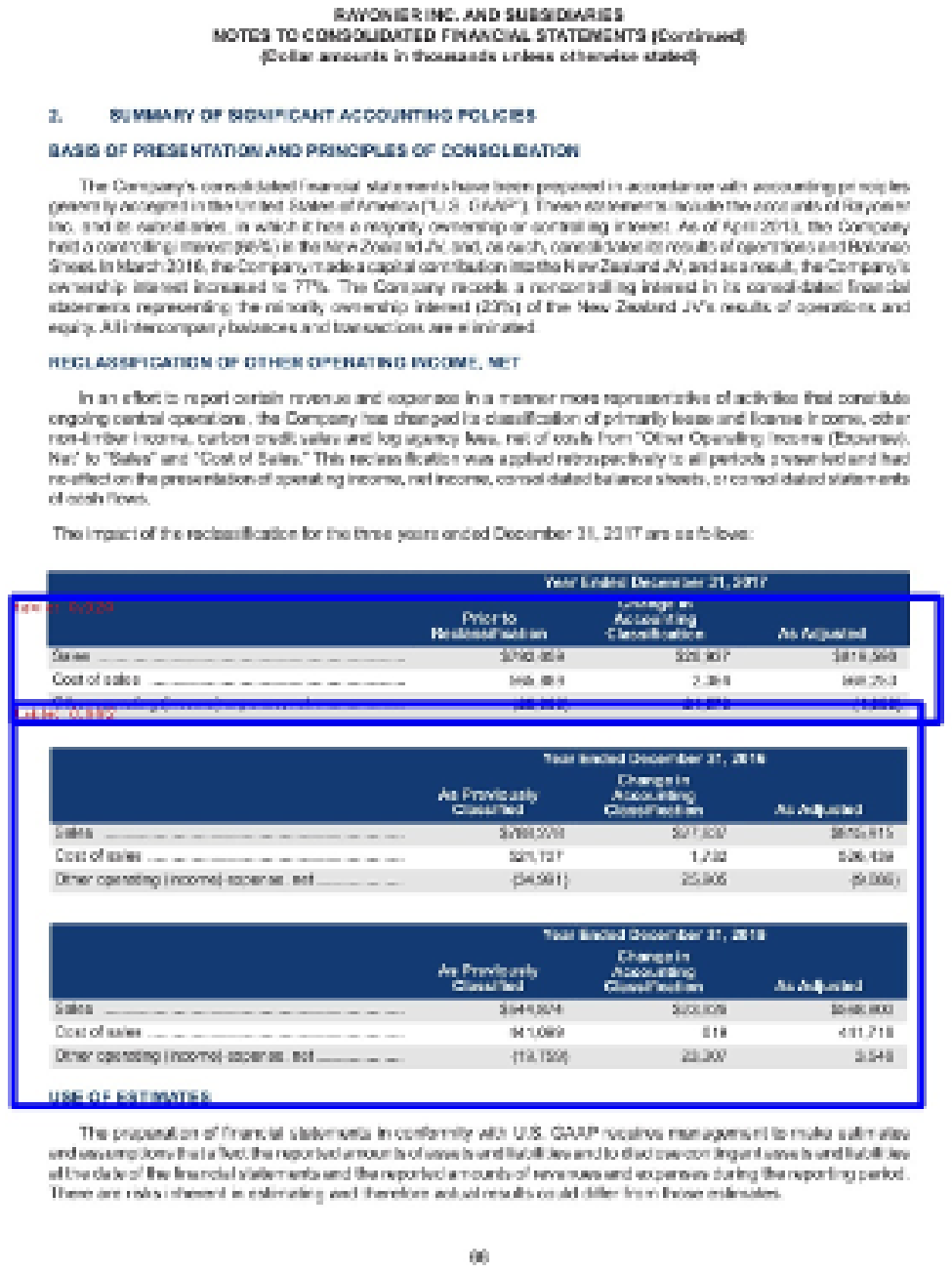, width=0.24\textwidth,height=0.18\textwidth}}
\hspace{-0.01\textwidth}
\tcbox[sharp corners, size = tight, boxrule=0.2mm, colframe=black, colback=white]{
\psfig{figure=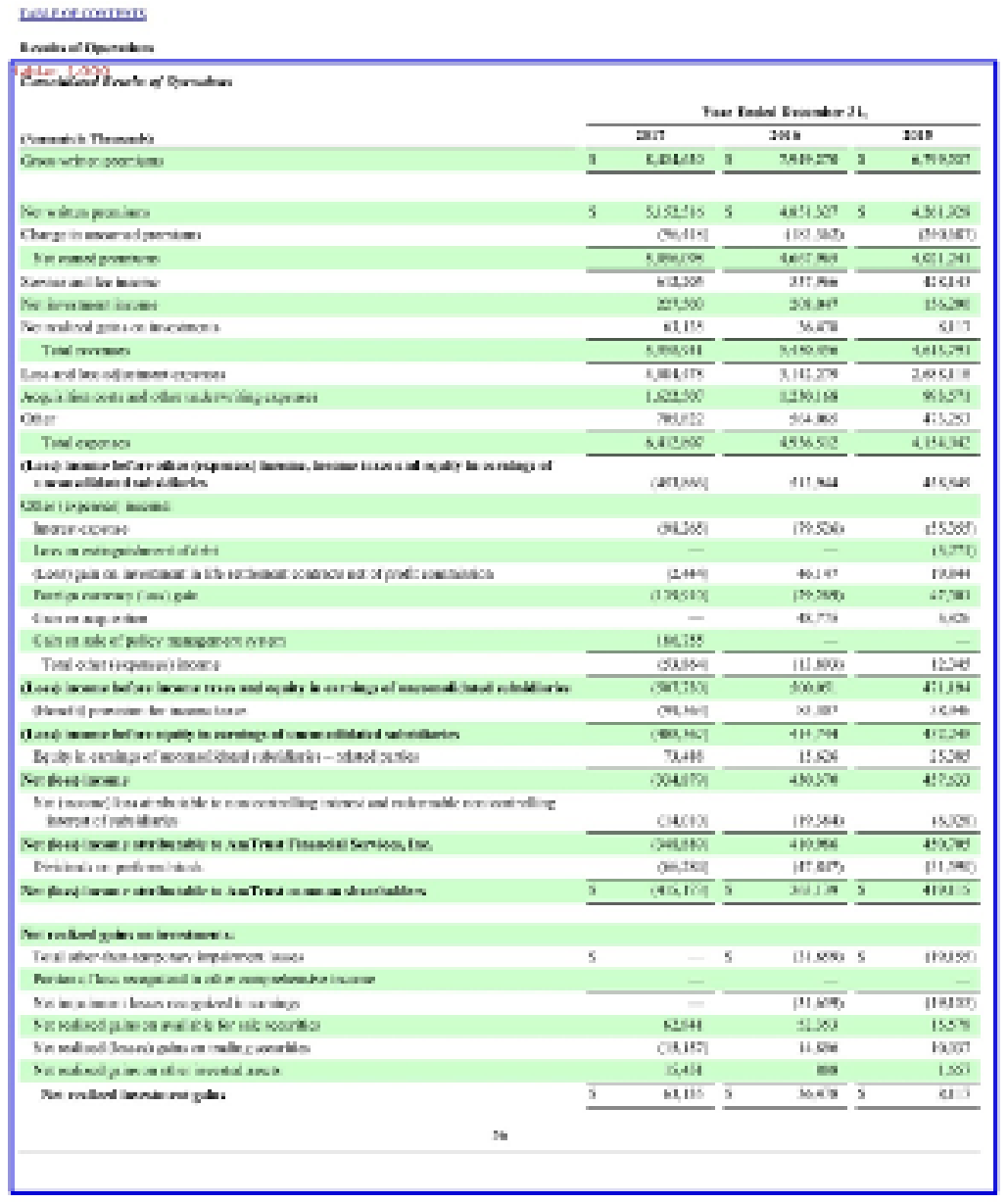, width=0.24\textwidth,height=0.18\textwidth}}}
\vspace{-0.005\textwidth}
\centerline{
\tcbox[sharp corners, size = tight, boxrule=0.2mm, colframe=black, colback=white]{
\psfig{figure=new_unlv/9508_067.eps, width=0.24\textwidth,height=0.18\textwidth}}
\hspace{-0.01\textwidth}
\tcbox[sharp corners, size = tight, boxrule=0.2mm, colframe=black, colback=white]{
\psfig{figure=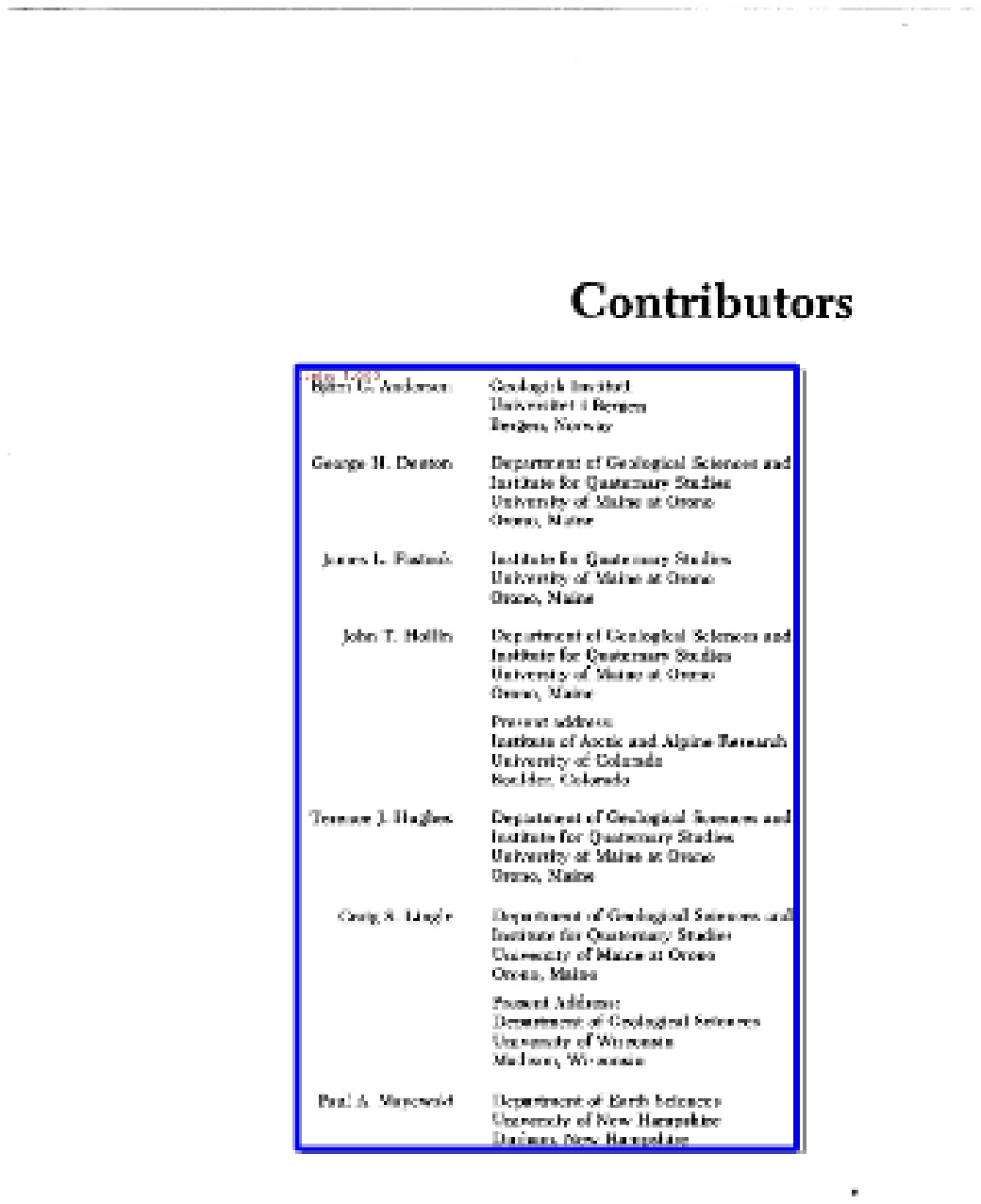, width=0.24\textwidth,height=0.18\textwidth}}
\hspace{0.01\textwidth}
\tcbox[sharp corners, size = tight, boxrule=0.2mm, colframe=black, colback=white]{
\psfig{figure=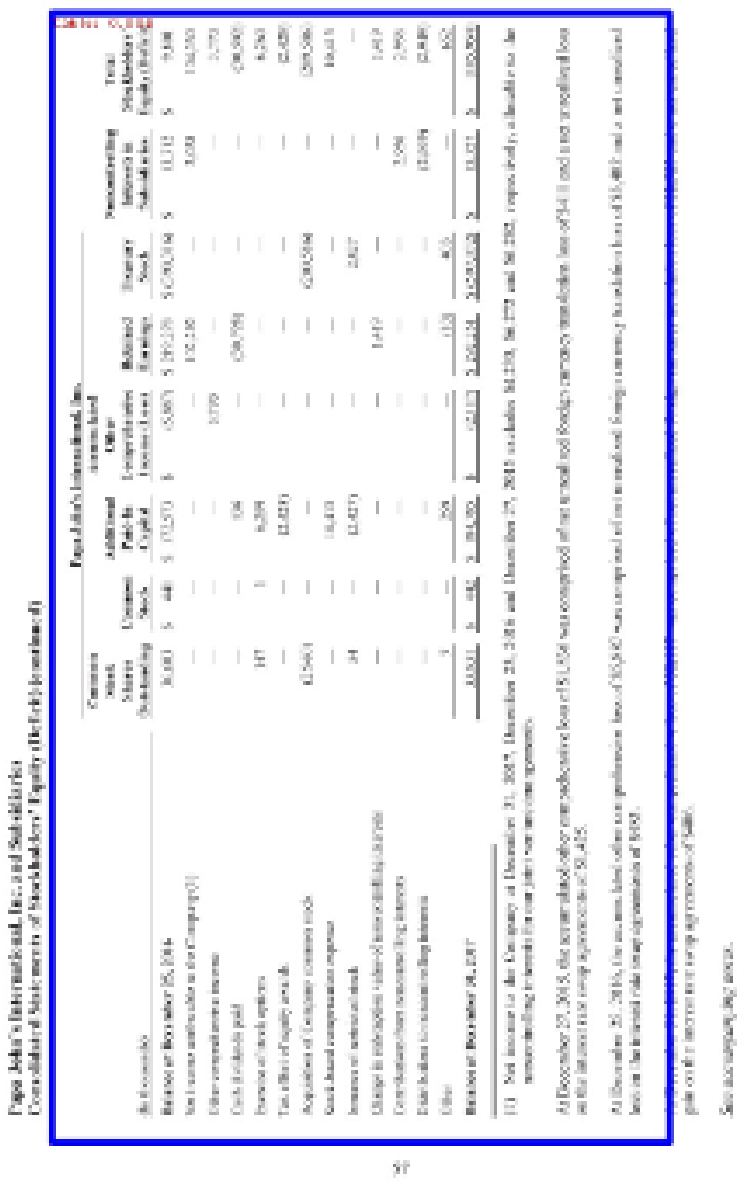, width=0.24\textwidth,height=0.18\textwidth}}
\hspace{-0.01\textwidth}
\tcbox[sharp corners, size = tight, boxrule=0.2mm, colframe=black, colback=white]{
\psfig{figure=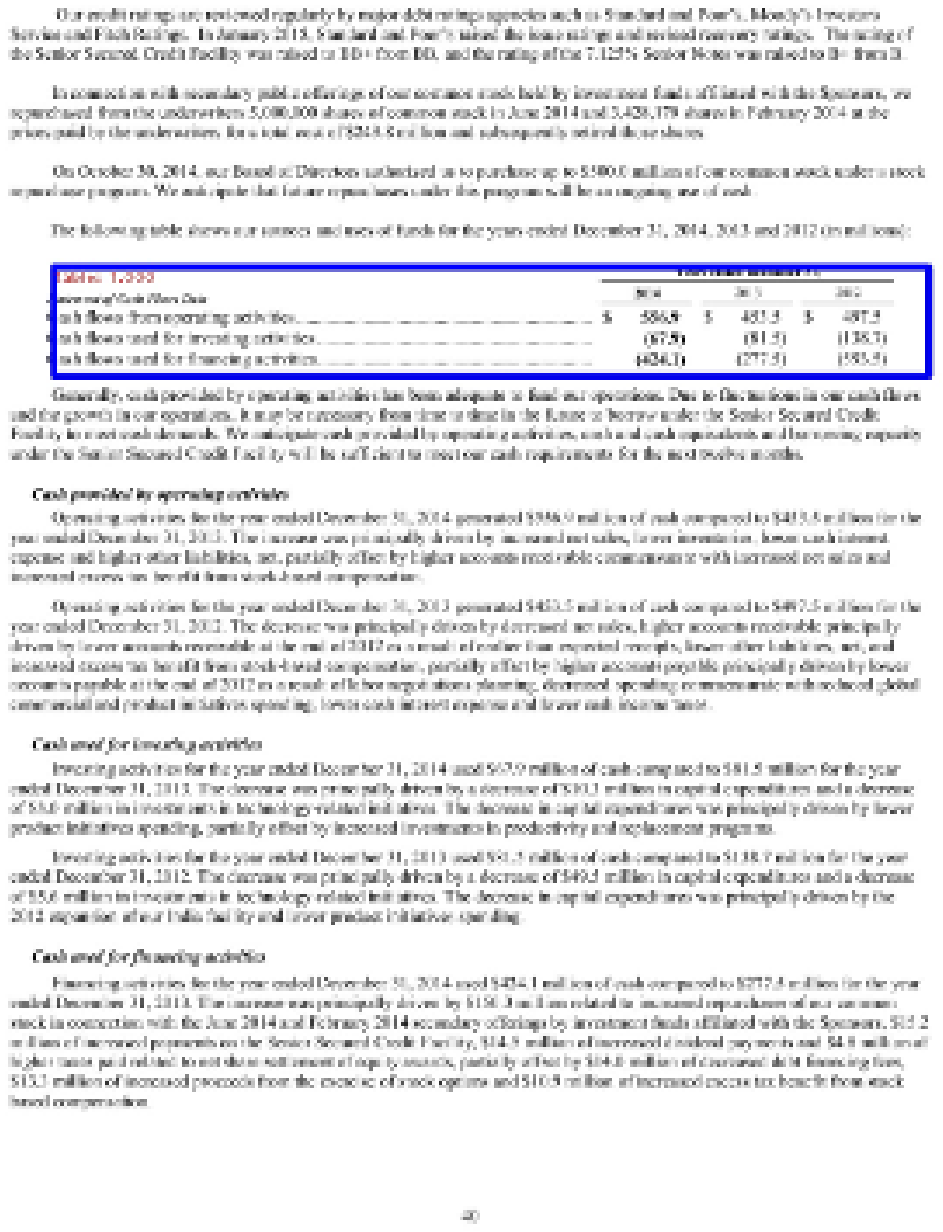, width=0.24\textwidth,height=0.18\textwidth}}}
\centerline{(a) \hspace{0.48\textwidth} (b)} 
\caption{(a) Results of the table localization using {\sc god} (Mask {\sc r-cnn}) in the document images of {\sc \textbf{unlv}} data set. (b) Results of table localization using the \textsc{god} (Mask {\sc r-cnn}) in the document images of {\sc\textbf{go-iiit-5k}} data set. Blue color represents the predicted bounding box of the table.\label{fig:result_unlv}}
\end{figure*}

\paragraph{\textbf{Comparison with the state-of-the-arts on }{\sc \textbf{icdar-2013}}}

Due to limited number of images, we consider this data set only for testing purpose. We use \textbf{Marmot} data set for training our model similar to DeepDeSRT~\cite{schreiber2017deepdesrt}. The \textsc{god} is compared to state-of-the-arts: Kavasidis {\em et al.}~\cite{kavasidis2018saliency}, DeepDeSRT~\cite{schreiber2017deepdesrt} and Tran {\em et al.}~\cite{tran2015table} on {\sc \textbf{icdar-2013}} table competition data set. Table~\ref{table_icdar_2013_ap} shows the comparison statistics. From the Table, it is observed that the \textsc{god} (Mask {\sc r-cnn}) and \textsc{god} (Faster {\sc r-cnn}) both obtained reasonably higher accuracy than the existing algorithms. Figure~\ref{fig:result_icdar2017}(b) displays the visual results obtained using the \textsc{god} (Mask {\sc r-cnn}). It can be found that the \textsc{god} (Mask {\sc r-cnn}) accurately detects the various types of tables with respect to style, content and size. This observation concludes that the \textsc{god} is robust with respect to variability of tables present in the document images.       
\begin{table}[ht!]
\begin{center}
\begin{tabular}{|l | l l l|} \hline
\textbf{Methods} & \multicolumn{3}{l|}{\textbf{Test Performance}} \\\cline{2-4} 
  & \textbf{Recall} & \textbf{Precision} & \textbf{F1} \\\hline  
Tran {\em et al.}~\cite{tran2015table} &0.964  &0.952 &0.958 \\
DeepDeSRT~\cite{schreiber2017deepdesrt} &0.962 &0.974 &0.968  \\
Kavasidis {\em et al.}~\cite{kavasidis2018saliency} &0.981 &0.975 &0.978 \\
\textsc{god}(Faster {\sc r-cnn})  &0.974  &\textbf{0.987} &0.981 \\ 
\textsc{god}(Mask {\sc r-cnn}) &\textbf{1.0} &0.982 &\textbf{0.991} \\ \hline
\end{tabular}
\end{center}
\caption{Comparison with state-of-the-arts based on recall, precision and F1 with IoU $=0.5$ on \textsc{\textbf{icdar-2013}} data set. Bold value indicates the best result.\label{table_icdar_2013_ap}} 
\end{table}

\paragraph{\textbf{Comparison with the state-of-the-arts on }{\sc \textbf{unlv}}}

We compare the performance of the \textsc{god} with the state-of-the-arts: Tesseract~\cite{smith2007overview}, Abbyy\footnote{https://www.
abbyy.com/en-eu/ocr-sdk/} and Gilani {\em et al.}~\cite{gilani2017table} on {\sc \textbf{unlv}} data set. Table~\ref{table_unlv_data} displays the comparison results. Figure~\ref{fig:result_unlv}(a) displays the results of table localization in document images on {\sc \textbf{unlv}} data set. Figure~\ref{fig:result_unlv} highlights that the \textsc{god} (Mask {\sc r-cnn}) is able to localize multiple tables with varying style in a single page.    
\begin{table}[ht!]
\begin{center}
\begin{tabular}{|l | l l l|} \hline
\textbf{Methods} & \multicolumn{3}{l|}{\textbf{Test Performance}} \\\cline{2-4} 
  & \textbf{Recall} & \textbf{Precision} & \textbf{F1} \\\hline  
Tesseract~\cite{smith2007overview} &0.643 &0.932 &0.761 \\
Abbyy                              &0.643 &0.950 &0.767  \\
Gilani {\em et al.}~\cite{gilani2017table} &0.907 &0.823 & 0.863  \\
\textsc{god}(Faster {\sc r-cnn}) &0.867 &0.929 &0.897  \\   
\textsc{god}(Mask {\sc r-cnn}) &\textbf{0.910} &\textbf{0.946} &\textbf{0.928}  \\ \hline
\end{tabular}
\end{center}
\caption{Comparison with state-of-the-arts on {\sc \textbf{unlv}} data set based on recall, precision and F1 with IoU $=0.5$. Bold value indicates the best result.\label{table_unlv_data}}
\end{table}

\subsection{Results on {\sc\textbf{go-}}{\sc \textbf{iiit}}\textbf{-5K} data set }
\begin{table}[ht!]
\begin{center}
\begin{tabular}{|l | l l l|l|} \hline
\textbf{Methods} & \multicolumn{4}{l|}{\textbf{Test Performance}} \\\cline{2-5} 
  & \textbf{Recall} & \textbf{Precision} & \textbf{F1} & \textbf{AP} \\\hline  
$^{*}$\textsc{god} (Faster {\sc r-cnn}) &0.8198  &0.8966 &0.8562 &0.8461 \\ 
$^{*}$\textsc{god} (Faster {\sc r-cnn})$^{\dagger}$ &0.8413 &0.9035 &0.8712 &0.8637 \\
$^{*}$\textsc{god} (Mask {\sc r-cnn}) &0.8995  &0.8562 &0.8778 &0.8734 \\ $^{*}$\textsc{god} (Mask {\sc r-cnn})$^{\dagger}$ &\textbf{0.9659}  &\textbf{0.9389} &\textbf{0.9524} & \textbf{0.9558} \\ \hline
$^{**}$\textsc{god} (Faster {\sc r-cnn}) &0.7538  &0.8425 &0.7936 &0.7632\\ 
$^{**}$\textsc{god} (Faster {\sc r-cnn})$^{\dagger}$ &0.7891 &0.8610 &0.8250 &0.7926 \\ 
$^{**}$\textsc{god} (Mask {\sc r-cnn}) &0.8841 &0.8510 &0.8676 &0.8567 \\ 
$^{**}$\textsc{god} (Mask {\sc r-cnn})$^{\dagger}$ &\textbf{0.9218} &\textbf{0.9283} &\textbf{0.9250} &\textbf{0.9199} \\ \hline
\end{tabular} 
\end{center}
\caption{Performance of the trained \textsc{god} model on {\sc\textbf{go-iiit-5k}} with IoU $=0.5$. {\sc \textbf{god:}} trained on public benchmark {\sc \textbf{icdar-2013}}, {\sc \textbf{icdar-pod2017}}, {\sc \textbf{unlv}} and \textbf{Marmot} data sets and tested on test images of {\sc\textbf{go-iiit-5k}} data set. {\sc \textbf{god$^{\dagger}$:}} trained on the public benchmark {\sc \textbf{icdar-2013}}, {\sc \textbf{icdar-pod2017}}, {\sc \textbf{unlv}} and \textbf{Marmot} data sets, then fine-tuned on training images and tested on test images of {\sc\textbf{go-iiit-5k}} data set. `*' indicates that data set is randomly divided into training and test sets. `**' indicates that data set is divided into training and test sets based on company. Bold value indicates the best result. \label{table_report_data}}
\end{table}

\begin{figure}[ht!]
\centerline{
\tcbox[sharp corners, size = tight, boxrule=0.2mm, colframe=black, colback=white]{
\psfig{figure=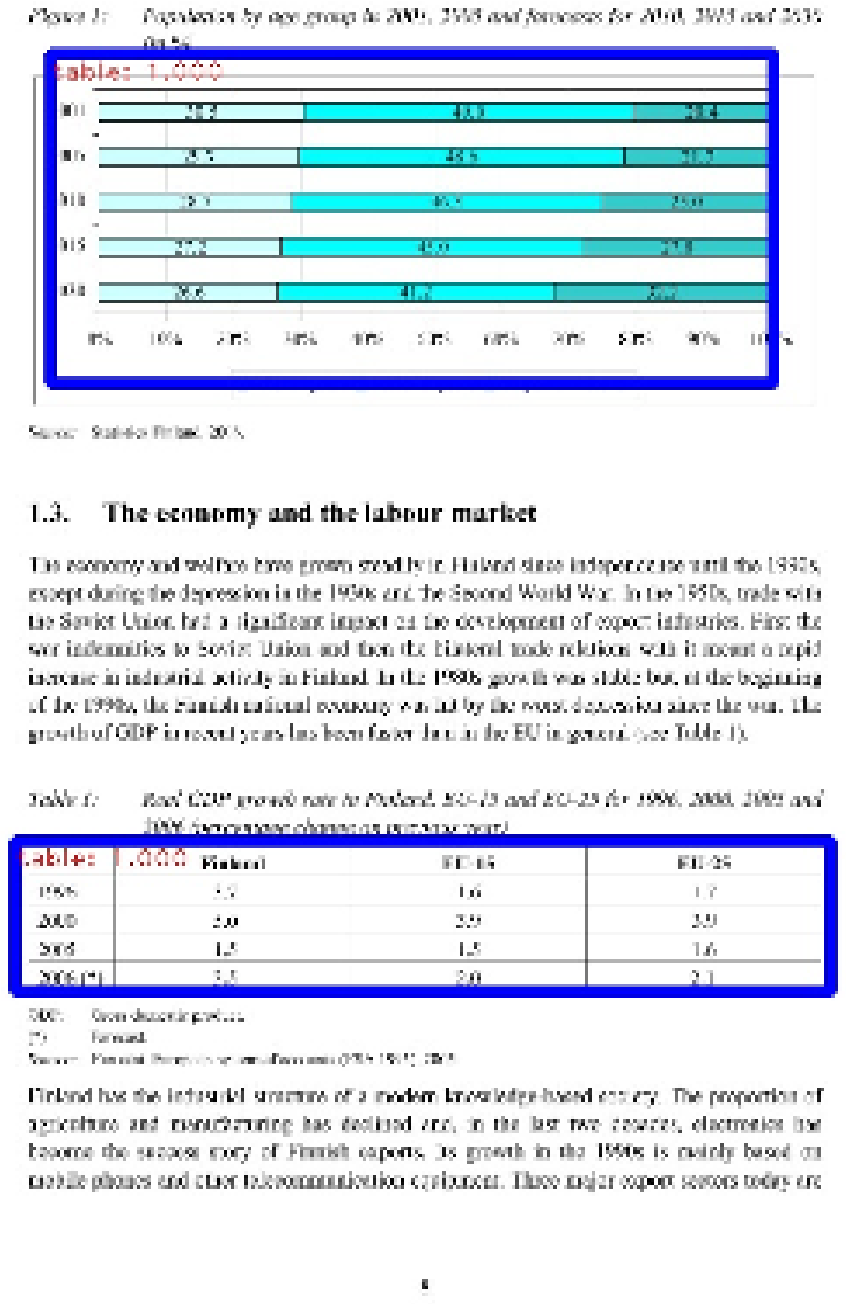, width=0.24\textwidth,height=0.18\textwidth}}
\hspace{-0.01\textwidth}
\tcbox[sharp corners, size = tight, boxrule=0.2mm, colframe=black, colback=white]{
\psfig{figure=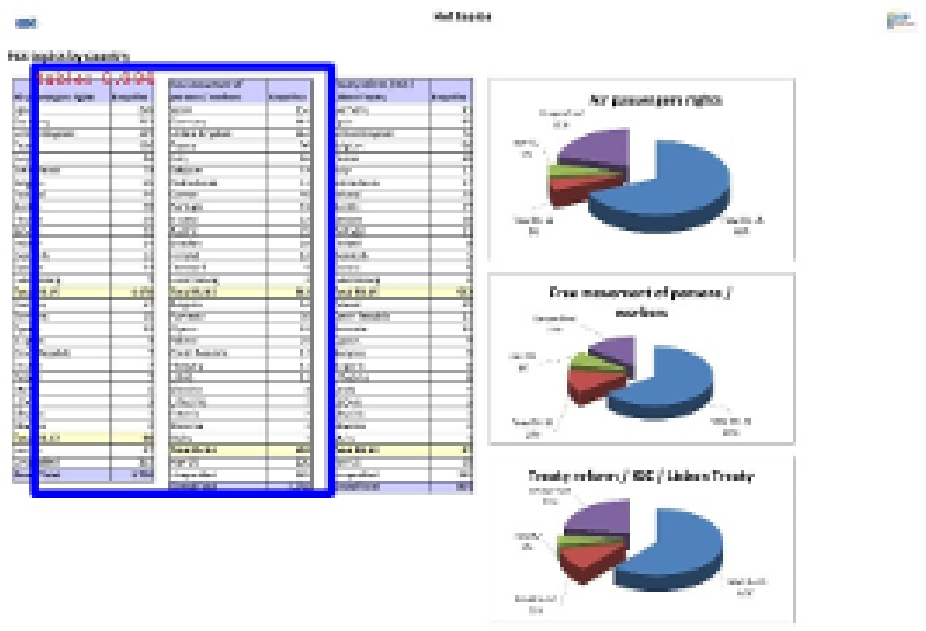, width=0.24\textwidth,height=0.18\textwidth}}}
\centerline{(a)\hspace{0.2\textwidth} (b)}
\vspace{0.003\textwidth}
\centerline{
\tcbox[sharp corners, size = tight, boxrule=0.2mm, colframe=black, colback=white]{
\psfig{figure=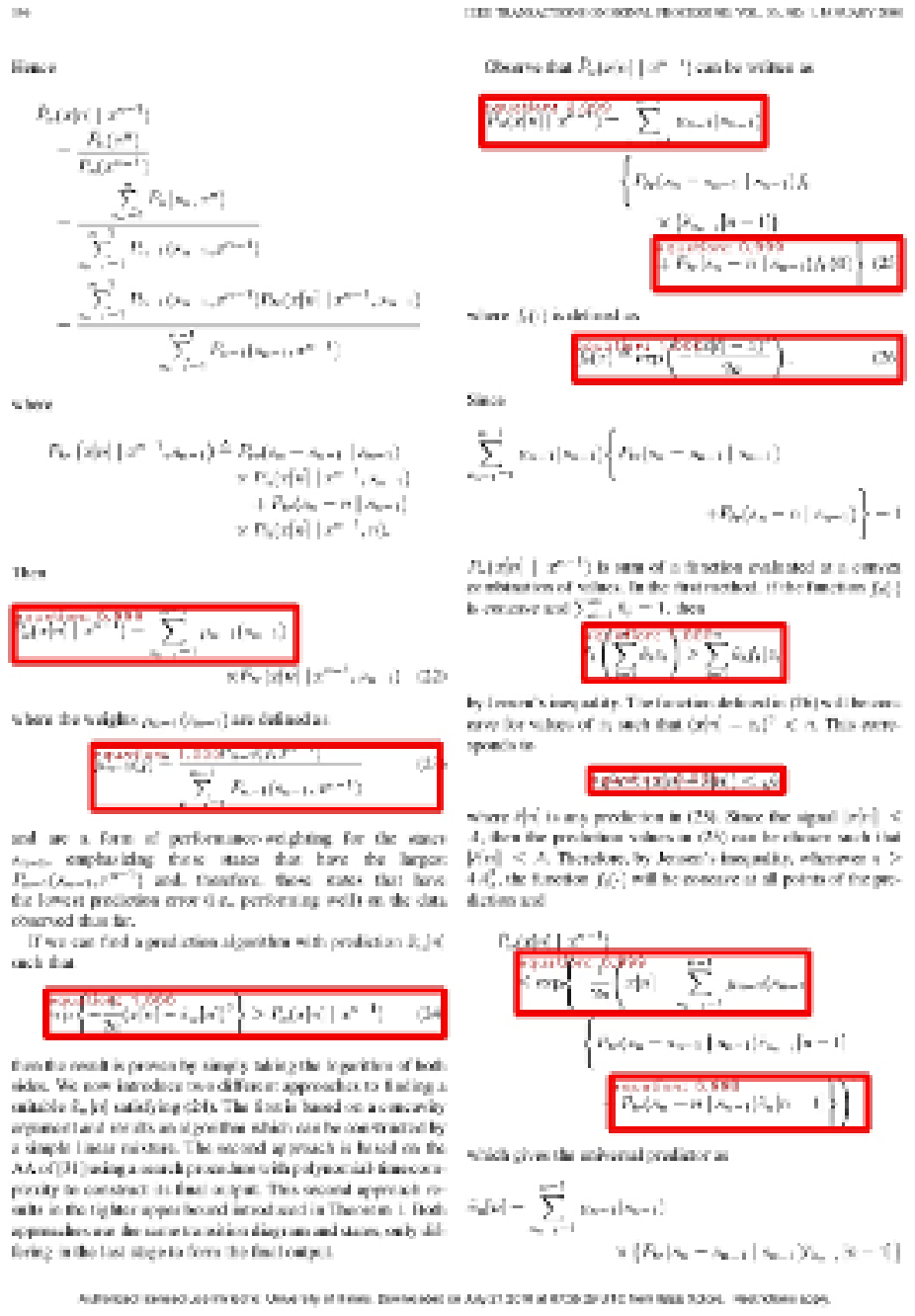, width=0.24\textwidth,height=0.18\textwidth}}
\hspace{-0.01\textwidth}
\tcbox[sharp corners, size = tight, boxrule=0.2mm, colframe=black, colback=white]{
\psfig{figure=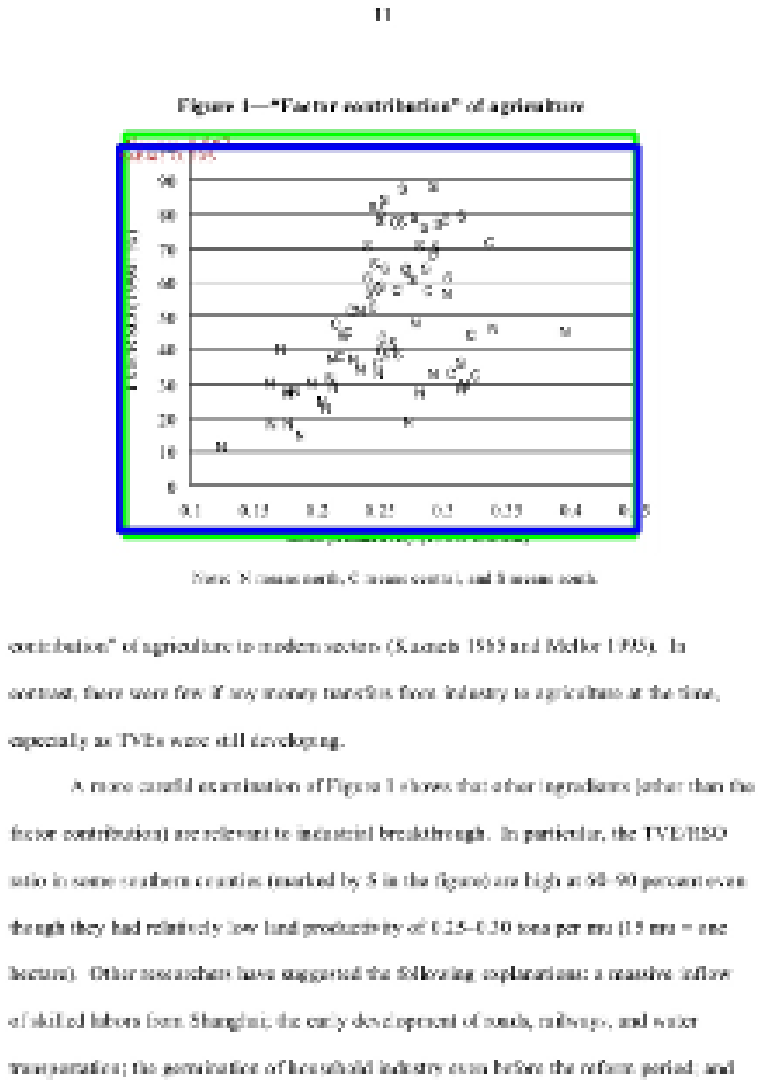, width=0.24\textwidth,height=0.18\textwidth}}}
\centerline{(c)\hspace{0.2\textwidth} (d)}
\vspace{0.003\textwidth}
\centerline{
\tcbox[sharp corners, size = tight, boxrule=0.2mm, colframe=black, colback=white]{
\psfig{figure=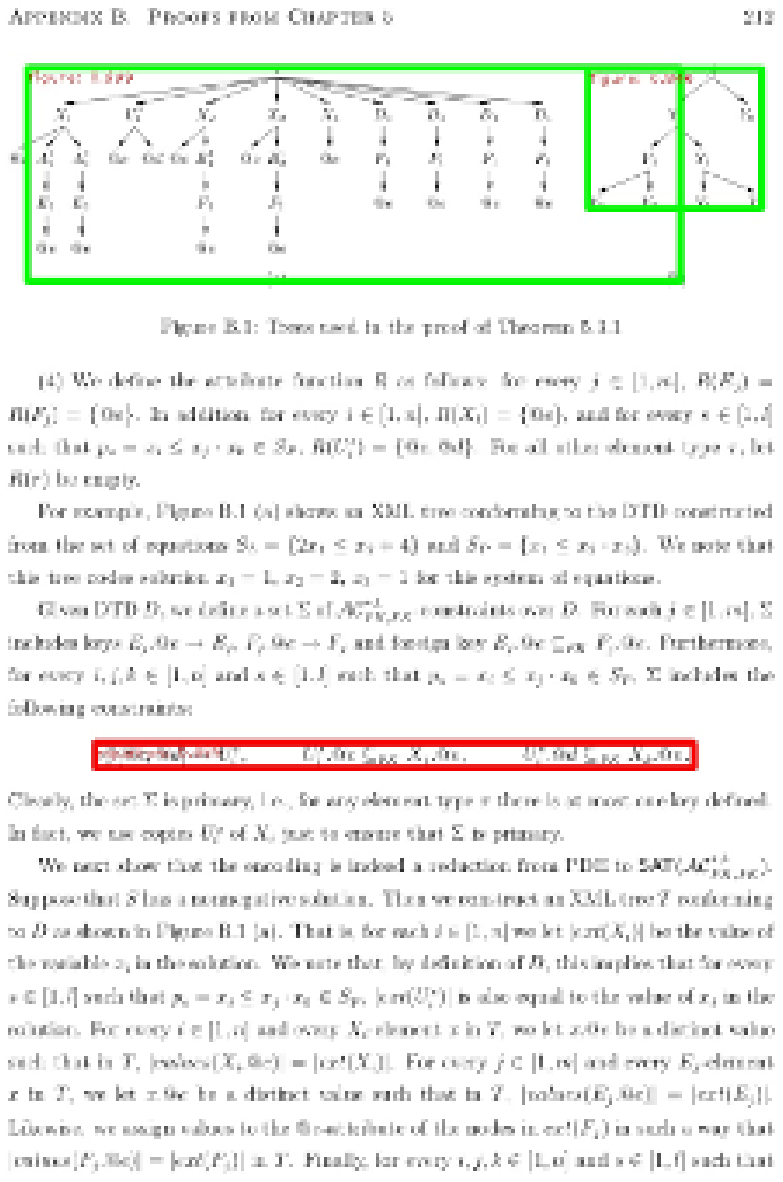, width=0.24\textwidth,height=0.18\textwidth}}
\hspace{-0.01\textwidth}
\tcbox[sharp corners, size = tight, boxrule=0.2mm, colframe=black, colback=white]{
\psfig{figure=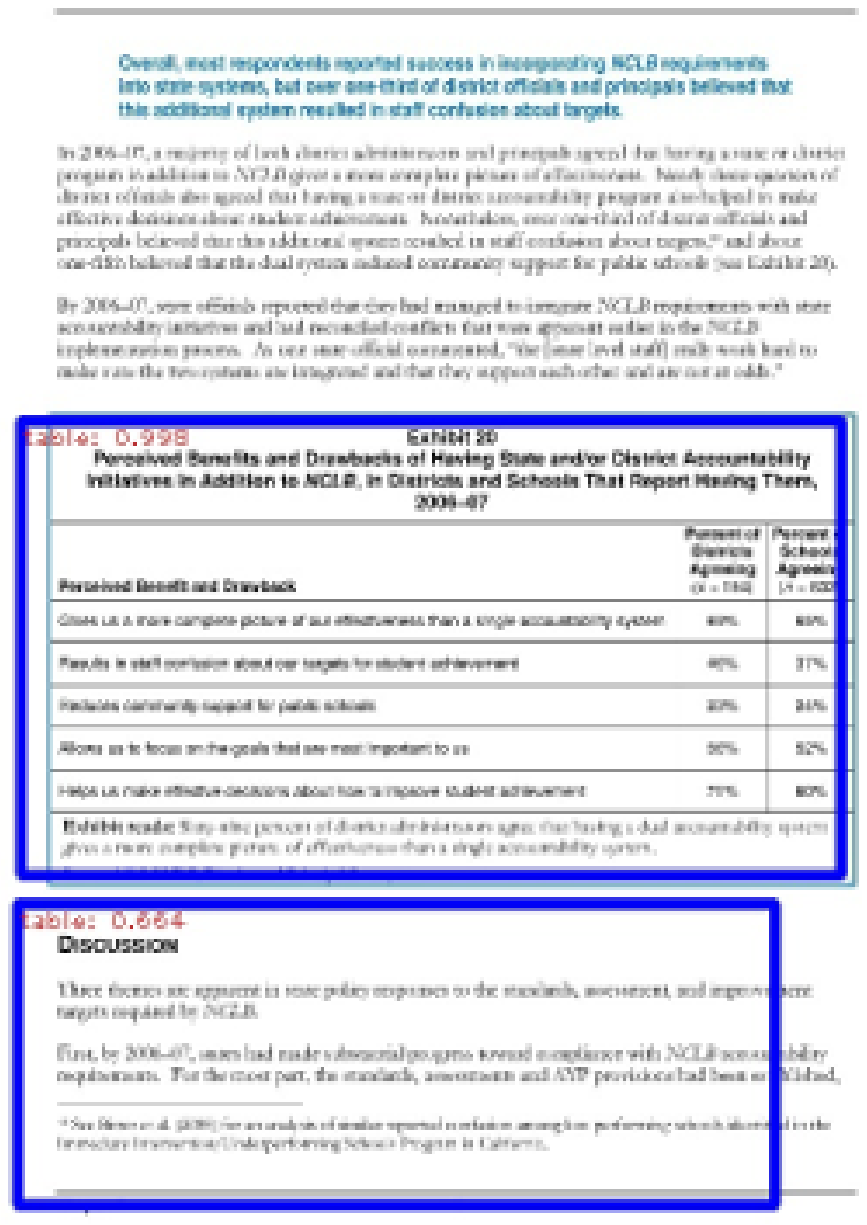, width=0.24\textwidth,height=0.18\textwidth}}}
\centerline{(e)\hspace{0.2\textwidth} (f)}
\caption{Examples where the \textsc{god} fails to accurately localize the graphical objects. (a) figure is detected as table. (b) table is partially detected (here, object of interest is table). (c) some equations are not detected. (d) figure is detected as both figure and table. (e) single figure is detected as multiple figures. (f) paragraph is detected as table. Blue, Green and Red colors represent the predicted bounding boxes of table, figure and equation, respectively. \label{fig:result_failure}}
\end{figure}

Though the \textsc{god} performs good on the public benchmark data sets. To establish the adaptability of the \textsc{god} trained model to localize tables in different types of documents, we show the results of table detection in documents of annual reports of various companies. For this purpose, we create a data set named as {\sc\textbf{go-iiit-5k}} by annotating $5$K pages of annual reports which are downloaded from \textit{ICAEW website}\footnote{\url{https://www.icaew.com/library/company-research/company-reports-and-profiles/annual-reports}}. This data set is divided into training set consisting of $3$K and test set consisting of $2$K images. This data set is different from {\sc \textbf{icadar-2013}}, {\sc \textbf{icdar-pod2017}} and {\sc \textbf{unlv}}. The trained \textsc{god} model on {\sc \textbf{icdar-2013}}, {\sc \textbf{icdar-pod2017}}, {\sc \textbf{unlv}} and \textbf{Marmot} data sets is tested on test images of {\sc\textbf{go-iiit-5k}} data set and obtain good accuracy. While we fine-tuned this trained model on training images of {\sc\textbf{go-iiit-5k}} data set and tested on test images of this data set, we obtain better table detection accuracy. It is also noted that \textsc{god} (Mask {\sc r-cnn}) performs better than {\sc god} (Faster {\sc r-cnn}) for both the cases: with out fine-tuning and with fine-tuning by training images of {\sc\textbf{go-iiit-5k}} data set. Table~\ref{table_report_data} highlights the performance of the \textsc{god} on {\sc\textbf{go-iiit-5k}} data set. Figure~\ref{fig:result_unlv}(b) shows the visual results obtained by {\sc god} (Mask {\sc r-cnn}). This experiment illustrates the adaptability of the trained \textsc{god} (Mask {\sc r-cnn}) model to localize the graphical objects mostly table in the different type of document images.  

The \textsc{god} method detects a wide range of graphical objects present in the documents giving very promising results. While the other existing methods need to object wise separate classifiers, the same \textsc{god} model works for the various categories of the graphical object. Detecting tables in the scientific document such as {\sc \textbf{icdar}} and {\sc \textbf{unlv}} data sets is an easy task as the tables or cells are well structured. The table detection task outperforms all the state-of-the-art methods in every aspect. Tables without well-defined boundaries of tables or cells are detected quite accurately. Detection of tables in the annual reports of various companies seem to be a difficult task since the format of the table varies a lot for different companies and some tables are not exactly aligned as a structured one; some tables are containing texts in between two cells, for which human opinion about whether the whole table should be considered as a table is also not uniform. Figures are generally easy to identify but they can also sometimes be tricky as some figures may contain some regions which are resembled to be the tables. Another difficult object to detect is equations as they are continued to the next few lines which sometimes are unaligned. Sometimes, equations are embedded in the text regions. When, the same types/categories of objects are in very close proximity and the objects are small in size such as equations as compared to the whole document, the model fails to accurately detect the regions between two such object regions to fall into any one of the object regions hence resulting in low m\textsc{AP} with high IoU score. But if the gap between two smaller sized objects is enough then objects of all sizes are localized pretty accurately. Our model outperforms the existing methods where objects were detected based on their fixed general structures of the objects.

Although, the \textsc{god} performs reasonably good to localize graphical objects: tables, figures and equations in the documents. However,
it sometime fails to localize those objects having ambiguity in appearance. Figure~\ref{fig:result_failure} displays some cases where the \textsc{god} fails to properly localize graphical objects present in the document images. From Figure~\ref{fig:result_failure} (a), it is observed that one graph with table type structure is detected as table. Multiple tables close to each other in the document, \textsc{god} detects them partially as a single table (Figure~\ref{fig:result_failure} (b)). Among multiple equations in the document, only few of them are detected by the \textsc{god} (Figure~\ref{fig:result_failure} (c)). One graph plot with vertical lines is detected as both table and figure (Figure~\ref{fig:result_failure} (d)). One figure in the document is detected as two different graphs (Figure~\ref{fig:result_failure} (e)). A paragraph is detected as table (Figure~\ref{fig:result_failure} (f)).            
\section{Conclusion}

This paper presents a novel end-to-end trainable deep learning based framework to localize graphical objects in the document images, called as Graphical Object Detection ({\sc god}), inspired by recent object detection algorithms: Faster {\sc r-cnn}~\cite{ren2015faster} and Mask {\sc r-cnn}~\cite{he2017mask}. To handle scarcity of labelled training samples, the {\sc god} employs the concept to transfer learning to localize graphical object in the document images. Experiments on the various public benchmark data sets conclude that our model yields promising results as compared to state-of-art techniques. From the experiments, it is also concluded that it is able to localize multiple tables, equations and figures with large variability present in the documents. The trained model is easily adapted to localize tables in the different kinds of documents like annual reports of various companies.

\def\IEEEbibitemsep{0pt plus.5pt}
\bibliographystyle{IEEEtran}
\bibliography{reference}
\end{document}